\documentclass[letterpaper]{article} 
\usepackage{aaai2026}  
\usepackage{times}  
\usepackage{helvet}  
\usepackage{courier}  
\usepackage[hyphens]{url}  
\usepackage{graphicx} 
\usepackage{natbib}  
\usepackage{caption} 
\usepackage{amsmath}
\usepackage{subcaption}
\usepackage{booktabs}
\usepackage{multirow}
\usepackage{makecell}

\title{Mixup Model Merge: Enhancing Model Merging Performance through Randomized Linear Interpolation}

\author{
    Yue Zhou\textsuperscript{\rm 1} \quad
    Yi Chang\textsuperscript{\rm 1,2,3} \quad
    Yuan Wu\textsuperscript{\rm 1}\thanks{Corresponding author}
}
\affiliations {
    \textsuperscript{\rm 1}School of Artificial Intelligence, Jilin University\\
    \textsuperscript{\rm 2}Engineering Research Center of Knowledge-Driven Human-Machine Intelligence, MOE, China\\
    \textsuperscript{\rm 3}International Center of Future Science, Jilin University\\
    yuezhou24@mails.jlu.edu.cn, yichang@jlu.edu.cn, yuanwu@jlu.edu.cn \\
}

\begin{document}

\maketitle

\begin{abstract}
Model merging aims to integrate multiple task-specific models into a unified model that inherits the capabilities of the task-specific models, without additional training.
Existing model merging methods often lack consideration of the varying contribution ratios of different task-specific models to the final merged model. In this paper, we propose Mixup Model Merge (M$^3$), a simple yet effective method inspired by the randomized linear interpolation strategy from the Mixup data augmentation technique.
M$^3$ performs randomized linear interpolation in parameter space between two task-specific LLMs, where interpolation coefficients are sampled from a Beta distribution to explore diverse contribution ratios.
This controllable randomness allows M$^3$ to outperform standard equal-ratio merging by discovering better contribution ratio combinations.
Extensive experiments show that M$^3$ significantly (1) improves merged LLM performance across tasks, (2) enhances out-of-distribution and adversarial robustness, and (3) outperforms the positive effects of the sparsification method DARE on model merging and can be further combined with DARE to achieve superior results. By tuning the Beta distribution’s shape parameters, (4) M$^3$ balances exploration efficiency and diversity in contribution ratios. 
The code is provided in the supplementary materials.
\begin{links}
    \link{Code}{https://github.com/MLGroupJLU/MixupModelMerge}
\end{links}
\end{abstract}

\section{Introduction}
In the field of Natural Language Processing (NLP), the emergence of large language models (LLMs) \citep{brown2020language,touvron2023llama,openai2023gpt,chowdhery2023palm} represents a revolutionary breakthrough. With their remarkable capabilities, these models have demonstrated outstanding performance across various tasks \citep{jiao2023chatgpt,chang2024survey,nam2024using,xing2024designing,guo2024chbench}, significantly advancing NLP technologies. 

\begin{figure}[t]
  \centering

  \begin{subfigure}[b]{\columnwidth}
    \centering
    \includegraphics[width=\columnwidth]{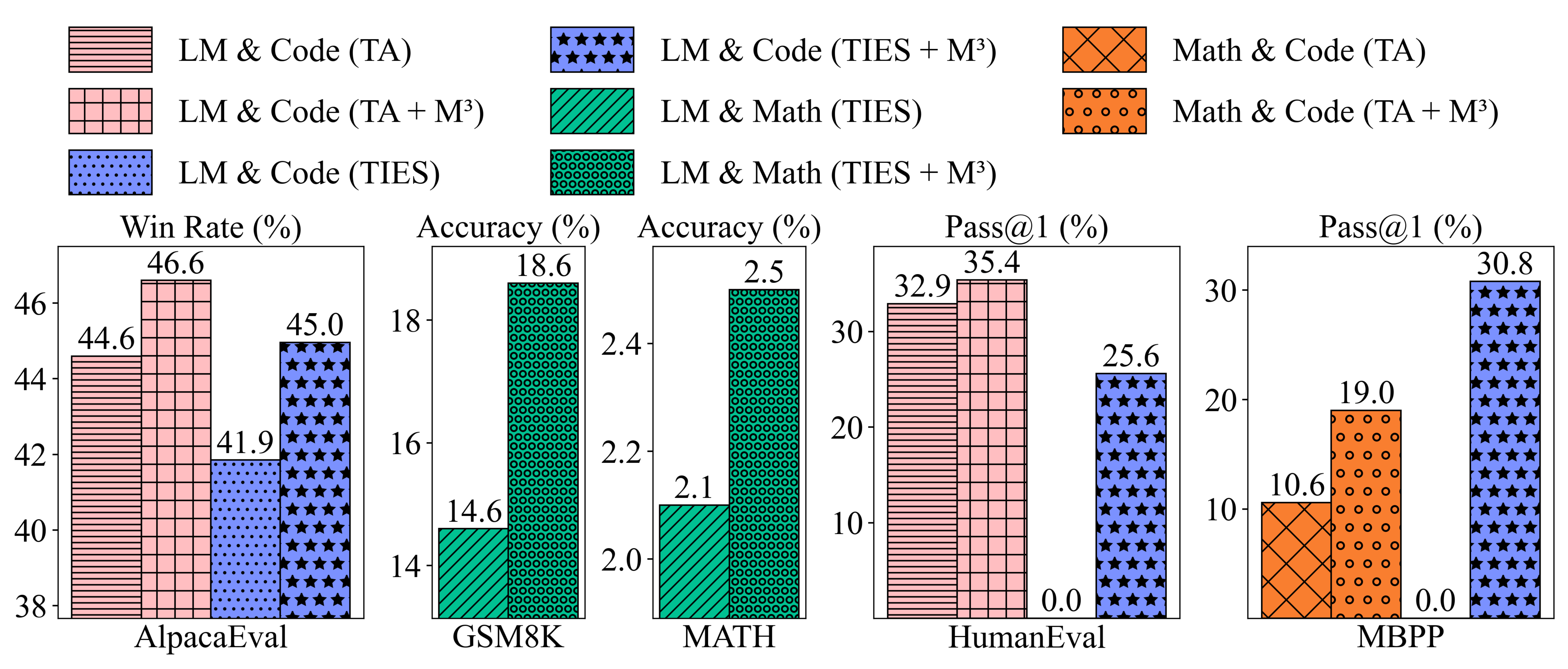}
    \caption{Performance of the merged models obtained through Task Arithmetic and TIES-Merging with or without M$^3$}
    \label{fig:introduction_1}
  \end{subfigure}

  \begin{subfigure}[b]{\columnwidth}
    \centering
    \includegraphics[width=\columnwidth]{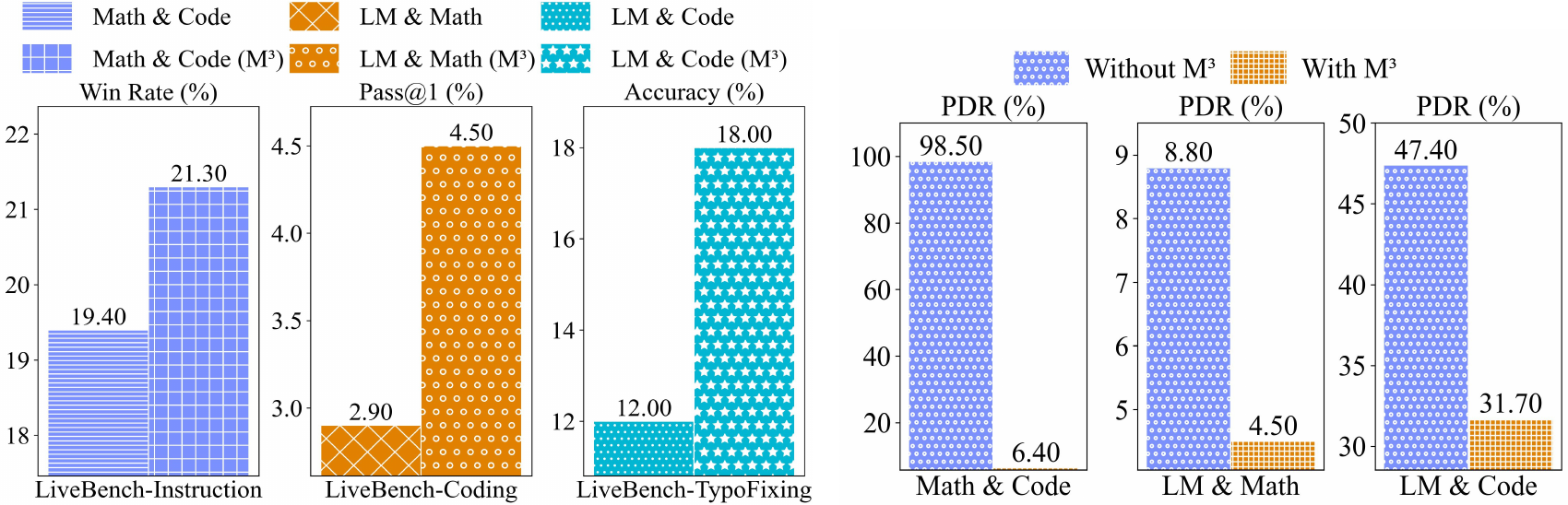}
    \caption{(Left) Performance on OOD datasets. (Right) Adversarial robustness (PDR) of merged models with or without M$^3$}
    \label{fig:introduction_2}
  \end{subfigure}

  \caption{(a) M$^3$ significantly improves the performance of model merging. (b) OOD and adversarial robustness of merged models with or without M$^3$}
  \label{fig:overall_comparison}
\end{figure}

\begin{figure*}[t]
  \centering
  \includegraphics[width=0.95\textwidth]{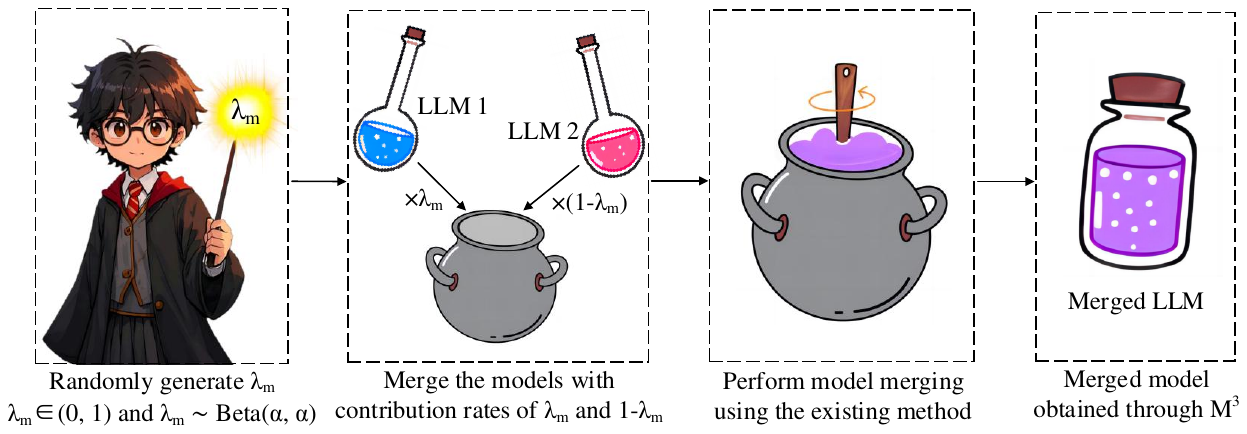}
  \caption{Implementation of M$^3$: A process analogous to proportionally mixing magical potions in Harry Potter. The proposed method controls the contribution ratio between two fine-tuned LLMs by randomly generating a linear interpolation ratio $\lambda_m$, where $\lambda_m \in (0, 1)$ and $\lambda_m \sim \text{Beta}(\alpha, \alpha)$. The distribution of $\lambda_m$ is controlled by adjusting $\alpha$.}
  \label{fig:HarryPotter}
\end{figure*}

Supervised Fine-Tuning (SFT) is a crucial technique for adapting LLMs to specific tasks, refining their performance by training on domain-specific data \citep{hu2021lora,ding2023parameter,xia2024rethinking}. However, SFT requires substantial computational resources and long training times \citep{brown2020language,chang2024ba}. To address this challenge, Model Merging has emerged as an efficient solution, fusing the parameters of multiple fine-tuned LLMs into a unified model with diverse capabilities, without the need for additional training or computational costs \citep{yang2024model,akiba2024evolutionary}. However, Existing model merging methods often adopt identical contribution ratios for different task-specific models in the merging process \citep{wortsman2022model,ilharco2022editing,matena2022merging,jin2022dataless,yadav2023resolving,yu2024language}, or rely on a few manually predefined contribution ratio values \citep{lin2023mitigating}. This restricts the exploration of how varying contribution ratios influence the final merged model, and limits the ability to fully explore the resulting parameter space. Consequently, the potential of the merged model is significantly constrained.

To address the above issue, we propose a novel approach, Mixup Model Merge (M$^3$), for exploring appropriate contribution ratios among task-specific models. This method is inspired by the randomized linear interpolation strategy from the Mixup data augmentation technique \citep{zhang2017mixup}. In contrast to conventional Mixup, M$^3$ performs linear interpolation in the model parameter space, where the interpolation coefficients are randomly sampled from a Beta distribution. Specifically, this method controls the contribution ratio between two fine-tuned LLMs by randomly generating a linear interpolation ratio $\lambda_m$, where $\lambda_m \in (0, 1)$ and $\lambda_m \sim \text{Beta}(\alpha, \alpha)$.
By adjusting the parameter $\alpha$, we can precisely control the distribution of $\lambda_m$, enabling M$^3$ to balance the diversity of $\lambda_m$ values with the efficiency of discovering suitable contribution ratios. In the parameter space, our method explores the family of models formed along the linear path between two task-specific LLMs by randomly sampling interpolation coefficients. This allows us to identify suitable interpolation coefficients (i.e., contribution ratios), thereby obtaining a merged model with superior performance.

We conducted extensive experiments with three homologous task-specific fine-tuned LLMs: WizardLM-13B~\citep{xu2024wizardlm}, WizardMath-13B~\citep{luo2023wizardmath}, and llama-2-13b-code-alpaca~\citep{chaudhary2023code}, which specialize in instruction following, mathematical reasoning, and code generation, respectively. 
Inspired by Mixup's effectiveness in enhancing the robustness of neural networks when handling corrupted labels or adversarial examples \citep{zhang2017mixup}, we further performed comprehensive evaluations on LiveBench \citep{white2024livebench} and PromptBench \citep{zhu2024promptbench} to validate the potential of M$^3$ in improving the OOD robustness and adversarial robustness \citep{wang2023robustness,zhu2023promptrobust} of merged models. The experimental results demonstrate that our proposed M$^3$ method can significantly improve merged models' performance across various tasks (as shown in Figure~\ref{fig:introduction_1}), and enhance the OOD and adversarial robustness of the merged models (as shown in Figure~\ref{fig:introduction_2}).

\section{Related Works}
\subsection{Model Merging}
Model merging is a technique that integrates the parameters of multiple models to create a unified model with enhanced or diverse capabilities \citep{wortsman2022model,ilharco2022editing,matena2022merging,jin2022dataless,yadav2023resolving,yu2024language,lin2024mitigating}.
Task arithmetic \citep{ilharco2022editing} leverages task vectors for model merging through arithmetic operations, incorporating a predefined scaling term to weight the contribution of different models.
Fisher Merging \citep{matena2022merging} performs parameter fusion by applying weights derived from the Fisher information matrix \citep{fisher1922mathematical}, resulting in more precise parameter integration.
TIES-Merging \citep{yadav2023resolving} addresses task conflicts by removing low-magnitude parameters, resolving sign disagreements, and merging only the parameters that align with the final agreed-upon sign.
In \citep{yu2024language}, it is found that LLMs can enhance their capabilities through model merging. Additionally, it introduces DARE, a method for sparsifying the delta parameters of the model \citep{ilharco2022editing}, significantly improving the performance of various model merging techniques. 
DELLA \citep{deep2024della} is a novel model merging technique that integrates a new pruning strategy called MAGPRUNE, which samples delta parameters based on their magnitudes. MAGPRUNE demonstrates improvements over existing methods such as DARE and TIES.

\section{Methodology}
We focus on merging the parameters of two homologous task-specific fine-tuned LLMs to construct a unified model that effectively preserves the capabilities of both, without requiring any further training.
 
\subsection{Mixup Model Merge}
Inspired by the random interpolation strategy of Mixup \citep{zhang2017mixup}, we extend the idea from input space to parameter space, proposing Mixup Model Merge (M$^3$).

We consider two task-specific fine-tuned language models, denoted as \(\theta_{t_1}^{\text{SFT}}\) and \(\theta_{t_2}^{\text{SFT}}\), obtained by independently fine-tuning a shared pretrained base model \(\theta_{\text{PRE}}\) on two different tasks \(t_1\) and \(t_2\), respectively. To explore the continuum of models between these two experts, we define a linear interpolation over their parameters:
\begin{equation}
\theta_M = \lambda_{m} \theta_{\text{SFT}}^{t_1} + (1 - \lambda_{m}) \theta_{\text{SFT}}^{t_2}, \quad \lambda_{m} \in (0, 1),
\label{eq:mixup_merge}
\end{equation}
This defines a one-dimensional family of interpolated models. Alternatively, the interpolation can be expressed in the \emph{delta space}, where we defined as the parameter offsets between the task-specific fine-tuned model and the pretrained model:
\begin{equation}
\delta_{t_i} = \theta_{\text{SFT}}^{t_i} - \theta_{\text{PRE}}, \quad i \in \{1, 2\},
\label{eq:delta_def}
\end{equation}
Then the interpolated model can be reformulated as:
\begin{equation}
\theta_M = \theta_{\text{PRE}} + \lambda_{m} \delta_{t_1} + (1 - \lambda_{m}) \delta_{t_2},
\label{eq:delta_form_merge}
\end{equation}
In both formulations, the resulting model \(\theta_M\) lies on the line segment connecting the two task-specific models in parameter space. This construction naturally defines a one-dimensional model family:
\begin{equation}
\mathcal{F}_M = \{ \theta_M^{\lambda_{m}} \mid \lambda_{m} \in (0, 1) \},
\label{eq:merge_family}
\end{equation}
Such a model family allows us to study how task behaviors interact and mix at the parameter level, and serves as the foundation for our proposed Mixup Model Merge ($M^3$).

\begin{figure}[t]
  \centering
  \includegraphics[width=0.9\columnwidth]
  {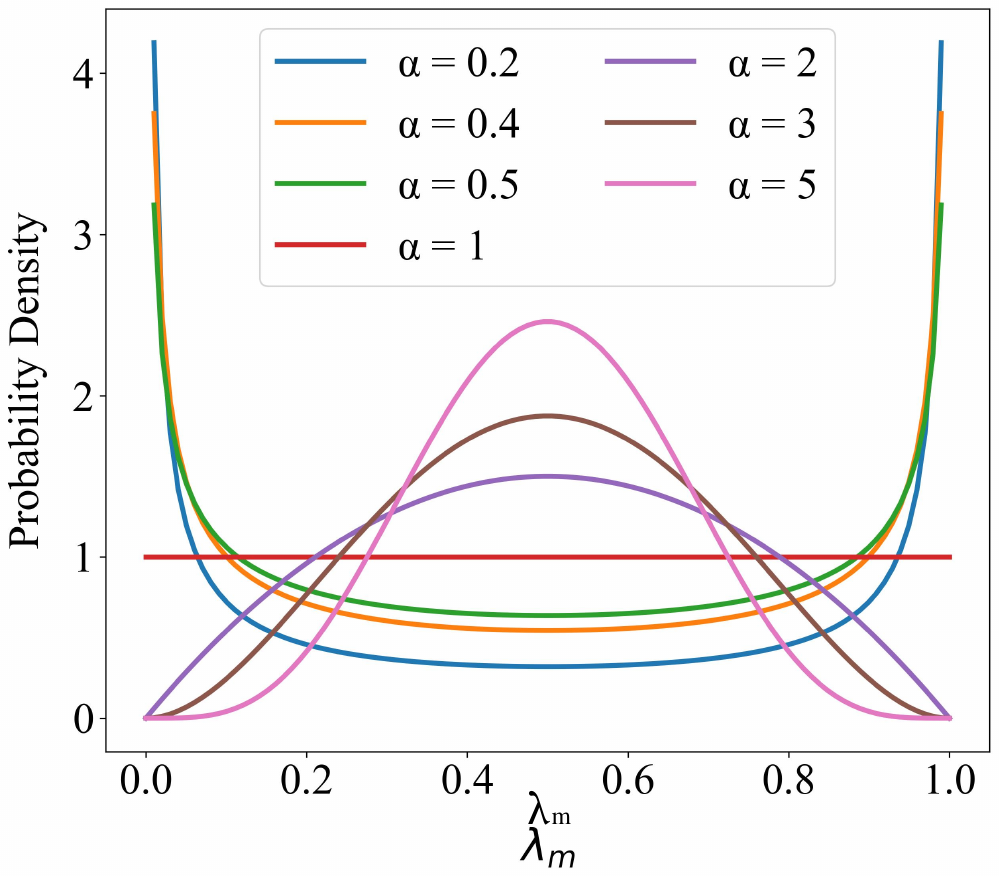}
  \caption{The Beta distribution visualization for different $\alpha$ values.}
  \label{fig:beta_distribution}
\end{figure}

\paragraph{Linear Interpolation Coefficient.}
A key component of the M$^3$ method is the stochastic sampling of the interpolation coefficient $\lambda_m$ from a Beta distribution, i.e., $\lambda_{m} \sim \text{Beta}(\alpha, \alpha)$. Defined over the interval $(0, 1)$, the Beta distribution is naturally suited for interpolation and introduces controllable randomness into the merging process. Its tunable shape parameter $\alpha$ allows us to adjust the sampling bias of $\lambda_m$, enabling targeted and efficient exploration of the one-dimensional model space $\mathcal{F}_M$ for optimal model combinations. Specifically, as illustrated in Figure \ref{fig:beta_distribution}:
\begin{itemize}
    \item When $\alpha < 1$, the distribution of $\lambda_{m}$ exhibits a bimodal shape, with higher probabilities near the extremes (0 and 1), indicating that the merged model is more likely to be dominated by one of the two task-specific models. As $\alpha$ decreases, the range of sampled $\lambda_{m}$ values becomes more concentrated, leading to higher exploration efficiency but lower diversity of the merged models explored.
    \item When $\alpha > 1$, the distribution of $\lambda_{m}$ is concentrated around the middle region (e.g., near 0.5), resulting in more balanced contributions from both models. As $\alpha$ increases, the range of sampled $\lambda_{m}$ values also becomes more concentrated, improving exploration efficiency while reducing the diversity of merged models.
    \item When $\alpha = 1$, $\lambda_{m}$ follows a uniform distribution, meaning that all values within the interval $(0, 1)$ are equally likely to be sampled. As $\alpha$ approaches 1, the distribution of $\lambda_{m}$ becomes more dispersed, leading to lower exploration efficiency but higher diversity in the merged models explored.
\end{itemize}



In Table \ref{table:sft_models_merging_results}, we report the specific $\lambda_m$ values that lead to performance improvements, along with the corresponding distribution shape parameter $\alpha$.


\paragraph{M$^3$ Implementation.}
M$^3$ is a plug-and-play approach that can be integrated into existing representative model merging methods to enhance the performance of the resulting merged models. As illustrative examples, we consider two widely used merging methods: Average Merging \citep{wortsman2022model} and Task Arithmetic \citep{ilharco2022editing}. 

The official computation process for Average Merging is described as follows:
\begin{equation}
\theta_M = \frac{1}{2} \left( \theta_{\text{SFT}}^{t_1} + \theta_{\text{SFT}}^{t_2} \right),
\end{equation}

The official computation process for Task Arithmetic is:
\begin{equation}
\begin{aligned}
\theta_M &= \theta_{\text{PRE}} + \lambda \cdot (\delta^{t_1} + \delta^{t_2}), \\
         &= \theta_{\text{PRE}} + \lambda \cdot \sum_{i=1}^2 (\theta_{\text{SFT}}^{t_i} - \theta_{\text{PRE}}),
\end{aligned}
\label{eq:merge_rewrite}
\end{equation}

$\lambda$ is a scaling term that takes values from the set $\{0.0, 0.1, \ldots, 1.0\}$. 

When introducing M$^3$, the process for Average Merging is reformulated as:
\begin{equation}
\theta_M = \lambda_m \theta_{\text{SFT}}^{t_1} + (1 - \lambda_m) \theta_{\text{SFT}}^{t_2},
\end{equation}
while the process for Task Arithmetic is reformulated as:
\begin{equation}
\theta_M = \theta_{\text{PRE}} + \lambda_m \delta^{t_1} + (1 - \lambda_m) \delta^{t_2},
\label{eq:delta_mix}
\end{equation}

where 
\begin{equation}
\lambda_m \in [0, 1], \quad \lambda_m \sim \mathrm{Beta}(\alpha, \alpha),
\label{eq:lambda_beta}
\end{equation}

$\lambda_m$ controls the interpolation ratio, i.e., the contribution ratio, between the two fine-tuned models. $\alpha$ determines the shape of the Beta distribution.

\begin{table*}[t]
\centering
\begin{tabular}{c c c c c c c c c c c} \toprule
\multirow{2}{*}{\makecell{Merging \\ Methods}} & 
\multirow{2}{*}{Models} & 
\multirow{2}{*}{\makecell{Use M$^3$ \\ (Ours)}} & 
\multicolumn{2}{c}{\makecell{Sampling\\ Details}} & 
\makecell{Instruction \\ Following} & 
\multicolumn{2}{c}{\makecell{Mathematical \\ Reasoning}} & 
\multicolumn{2}{c}{\makecell{Code Generating}} & 
\multirow{2}{*}{\textbf{Avg.}} \\ 
\cmidrule(lr){4-5} \cmidrule(lr){6-6} \cmidrule(lr){7-8} \cmidrule(lr){9-10}

& & &$\lambda_m$& $\alpha$ & AlpacaEval & GSM8K & MATH & HumanEval & MBPP &  \\ \hline

\multirow{3}{*}{-} & LM & - & -& - & 45.14& 2.20 & 0.04 & 36.59 & 34.00 & - \\ 
                   & Math & - & -& - & - & 64.22 & 14.02 & - & - & - \\ 
                   & Code & - & -& - & - & - & - & 23.78 & 27.60 & - \\ \hline
                   
\multirow{6}{*}{\makecell{Average \\ Merging}} & \multirow{2}{*}{\makecell{LM \\ \& Math}} & No & - & -& 45.28& 66.34& 13.40& - & - & 41.67 \\ 
                                                                         &                 & Yes& 0.35 & 2 & 44.40& 66.26& \underline{13.80}& - & - & 41.48 \\  \cline{2-11}

                                           & \multirow{2}{*}{\makecell{LM \\ \& Code}} & No & - & -&  36.60& -& -& 29.88& 32.00 & 32.82 \\  
                                                                         &             & Yes& 0.84 & 2 & \underline{43.91}$\uparrow$& -& -& \textbf{\underline{37.20}}$\uparrow$& \underline{34.40} & \underline{38.50}$\uparrow$ \\   \cline{2-11}

                                           & \multirow{2}{*}{\makecell{Math \\ \& Code}} & No& - & - & - & 56.17 & 10.28 & 8.53 & 8.20 & 20.80 \\  
                                                                           &             & Yes& 0.92 & 0.4 & - & \underline{63.61}$\uparrow$ & \underline{14.02}$\uparrow$& \underline{8.54} & \underline{19.20}$\uparrow$ & \underline{26.34}$\uparrow$ \\  \cline{1-11}

\multirow{6}{*}{\makecell{Task \\ Arithmetic}} & \multirow{2}{*}{\makecell{LM \\ \& Math}} & No & - & -& 45.78& 66.34& 13.40& - & - & 41.84 \\  
                                                                         &                 & Yes& 0.34 & 2 & 41.65& 66.34 & \underline{13.74}& -& - & 40.58 \\   \cline{2-11}

                                           & \multirow{2}{*}{\makecell{LM \\ \& Code}} & No & - & - & 44.64& - & - & 32.93& 33.60 & 37.06 \\  
                                                                         &             & Yes & 0.99 & 0.4 & \underline{46.64}& - & - & \underline{35.37} & \underline{33.80} & \underline{38.60} \\    \cline{2-11}

                                           & \multirow{2}{*}{\makecell{Math \\ \& Code}} & No & - & - & -& 64.67& 13.98& 8.54& 8.60 & 23.95 \\  
                                                                           &             & Yes & 0.98 & 0.5 & -& 63.53& 13.94& 7.93& \underline{19.00}$\uparrow$ & \underline{26.1} \\   \cline{1-11}

\multirow{6}{*}{\makecell{TIES-\\Merging}} & \multirow{2}{*}{\makecell{LM \\ \& Math}} & No & - & -& 38.63& 14.56& 2.12& - & - & 18.44 \\  
                                                                         &             & Yes& 0.62 & 1  & \underline{38.73}& \underline{18.57}$\uparrow$& \underline{2.48}& - & - & \underline{19.92} \\   \cline{2-11}
                                                                         
                                           & \multirow{2}{*}{\makecell{LM \\ \& Code}} & No & - & -&  41.85& -& -& 0.0 & 0.0 & 13.95\\  
                                                                         &             & Yes& 0.84 & 0.5 & \underline{44.96}& -& -& \underline{25.61}$\uparrow$ & \underline{30.80}$\uparrow$ & \underline{33.79}$\uparrow$\\   \cline{2-11}

                                           & \multirow{2}{*}{\makecell{Math \\ \& Code}} & No & - & -& - & 64.67 & 13.68 & 9.15 & 22.60 & 27.57 \\  
                                                                           &             & Yes& 0.63 & 3 & - & \underline{64.75} & \underline{14.16}& \underline{9.76} & 21.4 & 27.52 \\  \cline{1-11}
                                                                           
\multirow{6}{*}{DARE} & \multirow{2}{*}{\makecell{LM \\ \& Math}} & No & - & -& \textbf{49.00}& 66.64& 13.2& - & - & 42.95\\  
                                                              &   & Yes& 0.38 & 0.4  & 44.90& \textbf{\underline{67.32}}& \underline{13.74}& - & - & 41.98 \\   \cline{2-11}
                                                                         
                                           & \multirow{2}{*}{\makecell{LM \\ \& Code}} & No& - & - &  41.47& -& -& 35.98 & 33.00 & 36.82\\  
                                                                         &             & Yes& 0.99 & 0.5 & \underline{45.20}$\uparrow$& -& -& 35.98 & \textbf{\underline{35.20}} & \underline{38.79}\\   \cline{2-11}

                                           & \multirow{2}{*}{\makecell{Math \\ \& Code}} & No& - & - & - & 65.05 & 10.37 & 9.15 & 9.80 & 23.59 \\  
                                                                           &             & Yes& 0.98 & 0.5 & - & \underline{65.13} & \textbf{\underline{14.32}}$\uparrow$& 8.54 & \underline{18.00}$\uparrow$ & \underline{26.50}$\uparrow$ \\  \bottomrule

\end{tabular}
\caption{Comparison of Merged Model Performance Before and After Applying M$^3$. We use three task-specific LLMs: WizardLM-13B (LM), WizardMath-13B (Math), and llama-2-13b-codealpaca (Code). Bold indicates the best results for each dataset, underline highlights improvements with M$^3$, and $\uparrow$ denotes significant gains after introducing M$^3$.}
\label{table:sft_models_merging_results}
\end{table*}

\subsection{Theoretical Analysis}
\label{sec:Theoretical Analysis}
According to the theoretical and empirical findings of \citet{neyshabur2020being}, fine-tuning models from the same pretrained initialization can guide them into a shared, flat, and connected low-loss basin. As a result, the linear interpolation path between such fine-tuned models remains within this low-loss basin and does not encounter significant performance barriers. In some cases, models along the interpolation path even outperform the endpoints.

Since both $\theta_{t1}^{\mathrm{SFT}}$ and $\theta_{t2}^{\mathrm{SFT}}$ are fine-tuned from the same pretrained model, they can be considered to lie within the same flat and connected low-loss basin. Therefore, linearly interpolating between them is unlikely to cause a collapse in performance, and may even yield interpolated models with better performance than either endpoint. This provides a theoretical justification for exploring interpolated models along the linear path between two fine-tuned models as a means of finding superior merged models.

Consider the $i$-th parameter of the interpolated model:
\begin{equation}
\theta_{M,i} = \theta_{\mathrm{PRE},i} + \lambda_m \, \delta_{t1,i} + (1 - \lambda_m) \, \delta_{t2,i},
\end{equation}
where $\delta_{t1,i} = \theta_{t1,i}^{\mathrm{SFT}} - \theta_{\mathrm{PRE},i}$ and $\delta_{t2,i} = \theta_{t2,i}^{\mathrm{SFT}} - \theta_{\mathrm{PRE},i}$ denote the delta parameters from the pretrained model $\theta_{\mathrm{PRE}}$ to the task-specific fine-tuned models $\theta_{t1}^{\mathrm{SFT}}$ and $\theta_{t2}^{\mathrm{SFT}}$, respectively.

Define a function representing the change of the merged model’s $i$-th parameter relative to the pretrained model:
\begin{equation}
f_i(\lambda_m) = \lambda_m \, \delta_{t1,i} + (1 - \lambda_m) \, \delta_{t2,i},
\end{equation}
If $\delta_{t1,i} > 0$ and $\delta_{t2,i} < 0$, the delta parameters are in conflict and may partially cancel each other. In particular, when
\begin{equation}
\lambda_m = \frac{|\delta_{t2,i}|}{|\delta_{t1,i}| + |\delta_{t2,i}|} \in (0,1),
\end{equation}
we have $f_i(\lambda_m) = 0$, meaning the conflicting delta parameters are exactly canceled out at that parameter.

Even when $\lambda_m$ is close to this cancellation point, the magnitude of the update is significantly reduced. This shows that linear interpolation can effectively suppress conflicting delta parameters at certain parameters, providing stability in the merged model.

\section{Experiments}
\subsection{Experimental Setup}
\paragraph{Task-Specific Fine-Tuned LLMs and Datasets.}
Following the experimental setup given in \citet{yu2024language}, we select three task-specific fine-tuned LLMs: WizardLM-13B \citep{xu2024wizardlm}, WizardMath-13B \citep{luo2023wizardmath}, and llama-2-13b-code-alpaca \citep{chaudhary2023code}, all of which use Llama-2-13b \citep{touvron2023llama} as the pre-trained backbone. These models are respectively designed for instruction-following, mathematical reasoning, and code generation tasks.
To evaluate the instruction-following task we use AlpacaEval \citep{li2023alpacaeval}. For testing mathematical reasoning task, we employ GSM8K \citep{cobbe2021training} and MATH \citep{hendrycks2021measuring}. For estimating the performance of code-generating task, we use HumanEval \citep{chen2021evaluating} and MBPP \citep{austin2021program}. More details of these LLMs and datasets can be found in Appendix~A.

\paragraph{Benchmarks for evaluating Out-of-Distribution and Adversarial Robustness.}
To evaluate OOD robustness, we evaluate merged models including math \& code, LM \& math, and LM \& code on instruction following (LiveBench-Instruction), coding (LiveBench-Coding), and language comprehension (LiveBench-TypoFixing) category in LiveBench \citep{white2024livebench}, respectively. More details on OOD benchmarks are given in Appendix~B.

We utilize the Adversarial Prompt Attacks module in PromptBench \citep{zhu2024promptbench} to evaluate the robustness of LLMs against adversarial prompts. Specifically, we employ three attack methods: DeepWordBug (character-level) \citep{gao2018black}, BERTAttack (word-level) \citep{li2020bert}, and StressTest (sentence-level) \citep{naik2018stress}. The evaluation is conducted on two datasets supported by PromptBench: SST2 (sentiment analysis) \citep{socher2013recursive} and CoLA (grammatical correctness) \citep{warstadt2019neural}. For more details on PromptBench and attack methods, please refer to Appendix~C.

\paragraph{Evaluation Metrics.}
We calculate win rate for AlpacaEval and LiveBench-Instruction, zero-shot accuracy for GSM8K and MATH, pass@1 for HumanEval, MBPP and LiveBench-Coding, Matthews correlation coefficient (MCC) for CoLA, accuracy for SST2, and zero-shot accuracy for LiveBench-TypoFixing.

\paragraph{Implementation Details.}
In each distinct merging experiment---where ``distinct'' refers to variations in either the task-specific LLMs or the model merging methods---we perform a full sweep over the sole hyperparameter of M$^3$, $\alpha$, using the set $\{0.2, 0.4, 0.5, 1, 2, 3, 5\}$. For each $\alpha$ value, we sample \(\lambda_m\) once from the corresponding Beta distribution, resulting in a total of seven samples in each merging experiment. Seven samples are not the minimum required to observe significant performance gains. In most cases, a smaller number of samples is sufficient to identify a performant interpolation coefficient. However, to enable a more comprehensive exploration, we fix the number of samples to seven—each corresponding to a representative shape of the Beta distribution. While the stochastic nature of our method means there may be cases where none of the seven samples yields a significant improvement, such cases are not often observed in our experiments and thus not considered. Given the strong performance observed with just seven samples, and to conserve computational resources, we refrain from further sampling.

Unless otherwise specified, the remaining details of the model merging experiments follow \citet{yu2024language}. 
For a detailed description of the hyperparameter settings used in the representative existing model merging methods, please refer to Appendix~D.
All experiments are conducted on NVIDIA GeForce RTX 4090 GPUs.

\begin{table}[t]
\centering
\setlength{\tabcolsep}{1mm} 
\begin{tabular}{c c c c c c} \toprule
Model & Dataset & \makecell{Use\\ Mixup} & \makecell{Use \\Attack} & Metric (\%) & PDR (\%) \\ \hline

\multirow{8}{*}{\makecell{Math \\ \& \\ Code}} & \multirow{4}{*}{SST2} & \multirow{2}{*}{No} & No  & 57.68 & \multirow{2}{*}{38.97} \\  
                                               &                        &                     & Yes & 35.21 &                         \\ \cline{3-6}
                                               &                        & \multirow{2}{*}{Yes}& No  & \underline{86.24} & \multirow{2}{*}{\textbf{35.77}} \\  
                                               &                        &                     & Yes & 55.39 &                         \\ \cline{2-6}
                                               & \multirow{4}{*}{CoLA} & \multirow{2}{*}{No} & No  & 45.54 & \multirow{2}{*}{98.53} \\  
                                               &                        &                     & Yes & 0.67  &                         \\ \cline{3-6}
                                               &                        & \multirow{2}{*}{Yes}& No  & \underline{71.72} & \multirow{2}{*}{\textbf{6.42}} \\  
                                               &                        &                     & Yes & 67.11 &                         \\ \cline{1-6}

\multirow{8}{*}{\makecell{LM \\ \& \\ Math}} & \multirow{4}{*}{SST2} & \multirow{2}{*}{No}  & No  & 92.78 & \multirow{2}{*}{\textbf{29.05}} \\  
                                            &                        &                      & Yes & 65.83 &                         \\ \cline{3-6}
                                            &                        & \multirow{2}{*}{Yes} & No  & \underline{91.28} & \multirow{2}{*}{34.55} \\  
                                            &                        &                      & Yes & 59.75 &                         \\ \cline{2-6}
                                            & \multirow{4}{*}{CoLA}  & \multirow{2}{*}{No}  & No  & 79.19 & \multirow{2}{*}{8.84} \\  
                                            &                        &                      & Yes & 72.20 &                         \\ \cline{3-6}
                                            &                        & \multirow{2}{*}{Yes} & No  & \underline{80.54} & \multirow{2}{*}{\textbf{4.52}} \\  
                                            &                        &                      & Yes & 76.89 &                         \\ \cline{1-6}

\multirow{8}{*}{\makecell{LM \\ \& \\ Code}} & \multirow{4}{*}{SST2} & \multirow{2}{*}{No} & No  & 10.55 & \multirow{2}{*}{38.04} \\  
                                            &                        &                     & Yes & 6.54  &                         \\ \cline{3-6}
                                            &                        & \multirow{2}{*}{Yes}& No  & \underline{73.17} & \multirow{2}{*}{\textbf{7.68}} \\  
                                            &                        &                     & Yes & 67.55 &                         \\ \cline{2-6}
                                            & \multirow{4}{*}{CoLA} & \multirow{2}{*}{No} & No  & 74.21 & \multirow{2}{*}{47.42} \\  
                                            &                        &                     & Yes & 39.02 &                         \\ \cline{3-6}
                                            &                        & \multirow{2}{*}{Yes}& No  & \underline{74.78} & \multirow{2}{*}{\textbf{31.67}} \\  
                                            &                        &                     & Yes & 51.10 &                         \\ \bottomrule
\end{tabular}
\caption{Adversarial robustness of merged models on SST2 and CoLA with StressTest prompt attacks. Best and second-best results are marked in bold and underlined, respectively.}
\label{table:attack_results_stresstest}
\end{table}

\begin{table*}[t]
\centering
\begin{tabular}{c c c c c c c c c} \toprule
\multirow{2}{*}{\makecell{Merging \\ Methods}} & \multirow{2}{*}{Models} & \multirow{2}{*}{\makecell{Use M$^3$ \\ (Ours)}} & \multirow{2}{*}{\makecell{Use \\ DARE}} 
& \makecell{Instruction \\ Following} & \multicolumn{2}{c}{\makecell{Mathematical \\ Reasoning}} & \multicolumn{2}{c}{\makecell{Code Generating}} \\ \cmidrule(lr){5-5} \cmidrule(lr){6-7} \cmidrule(lr){8-9}
& & & & AlpacaEval & GSM8K & MATH & HumanEval & MBPP \\ \hline
\multirow{9}{*}{\makecell{Average \\ Merging}} & \multirow{3}{*}{\makecell{LM \\ \& Math}} & Yes & No & \textbf{44.40}& 66.26& \underline{13.80}& - & - \\
                                                                         &                 & No & Yes & \underline{44.22}& \textbf{66.57}& 12.96& -&-\\ 
                                                                         &                 & Yes & Yes & 43.53& \textbf{66.57}& \textbf{14.12}& -&-\\ \cline{2-9}

                                           & \multirow{3}{*}{\makecell{LM \\ \& Code}}  & Yes & No & \textbf{43.91}& -& -& \textbf{37.20}& \underline{34.40}\\
                                                                         &              & No & Yes & 38.81& -& -& 31.71& 32.40\\  
                                                                         &             & Yes & Yes & \underline{40.31}& -& -& \underline{36.59}&\textbf{37.00}\\ \cline{2-9}

                                           & \multirow{3}{*}{\makecell{Math \\ \& Code}}  & Yes & No & - & \underline{63.61} & \textbf{14.02} & \underline{8.54} & \underline{19.20}$\uparrow$ \\  
                                                                           &               & No & Yes & -& 56.18& 10.28& 6.10& 7.80\\    
                                                                           &             & Yes& Yes& -& \textbf{64.97}& \underline{13.54}& \textbf{9.76}& \textbf{21.20}\\ \cline{2-9}
                                                                           
\multirow{9}{*}{\makecell{Task \\ Arithmetic}} & \multirow{3}{*}{\makecell{LM \\ \& Math}}   & Yes & No & 41.65& 66.34 & \textbf{13.74}& -& -\\
                                                                                 &           & No & Yes & \textbf{49.00}& \underline{66.64}& 13.02& -& -\\  
                                                                                 &           & Yes & Yes & \underline{44.90} & \textbf{67.32}& \textbf{13.74}& -&-\\ \cline{2-9}

                                           & \multirow{3}{*}{\makecell{LM \\ \& Code}} & Yes &No & \textbf{46.64} & - & - & 35.37 & \underline{33.80} \\ 
                                                                         &             & No & Yes & 41.47 & -& -& \textbf{35.98}& 33.00\\ 
                                                                         &             & Yes & Yes & \underline{45.20} & -& -& \textbf{35.98}& \textbf{35.20}\\ \cline{2-9}

                                           & \multirow{3}{*}{\makecell{Math \\ \& Code}} & Yes &No & -& 63.53& 13.94& 7.93& \textbf{19.00}\\
                                                                           &             & No & Yes & -& \underline{65.05}& \underline{13.96}& \textbf{10.37}& 9.80\\  
                                                                           &             & Yes & Yes & -& \textbf{65.13}& \textbf{14.32}& \underline{8.54}& \underline{18.00}\\ \cline{2-9}
                                                                           
\multirow{9}{*}{\makecell{TIES-\\Merging}} & \multirow{3}{*}{\makecell{LM \\ \& Math}} & Yes & No & \underline{38.73}& \underline{18.57}& \underline{2.48}& - & - \\  
                                                                         &             & No & Yes & 37.92& 18.04& 2.34& - & - \\     
                                                                         &             & Yes & Yes & \textbf{39.93}& \textbf{19.26}& \textbf{2.82}& - & -\\ \cline{2-9}

                                           & \multirow{3}{*}{\makecell{LM \\ \& Code}} & Yes & No & \underline{44.96}& -& -& \underline{25.61}& \underline{30.80}\\  
                                                                         &             & No & Yes & 43.13& -& -& 0.0& 0.0\\       
                                                                         &             & Yes & Yes & \textbf{45.65}& -& -& \textbf{26.83}& \textbf{33.20}\\ \cline{2-9}

                                           & \multirow{3}{*}{\makecell{Math \\ \& Code}} & Yes & No & - & 64.75 & \underline{14.16} & \underline{9.76} & \underline{21.4} \\  
                                                                           &             & No & Yes & -& \textbf{64.82}& 13.88 & \textbf{10.37}& \textbf{23.60}\\ 
                                                                           &             & Yes& Yes& -& \underline{64.75}& \textbf{14.78} & 9.15& 19.60\\ \bottomrule
           
\end{tabular}
\caption{Comparison of our method M$^3$ and DARE. The best and second-best results are marked in bold and underlined fonts.
}
\label{table:DARE_M3}
\end{table*}

\subsection{Merging Task-Specific Fine-Tuned LLMs}
Our proposed method, M$^3$, is a plug-and-play approach. To evaluate its effectiveness in improving merged model performance, we integrate M$^3$ into several representative existing model merging methods (Average Merging, Task Arithmetic, TIES-Merging, DARE) and compare the performance of the merged models before and after applying M$^3$. The results is presented in Table \ref{table:sft_models_merging_results}.

\paragraph{Performance Gains of M$^3$ Across Merging Methods and Tasks.}
As shown in Table \ref{table:sft_models_merging_results}, M$^3$ generally enhances existing model merging methods, achieving over 10-point improvements on multiple datasets and up to 30 points on one.
For example, the improvements achieved by Average Merging with M$^3$ for Math \& Code are 7.43\% on GSM8K, 3.74\% on Math, and 11.0\% on MBPP. For LM \& Code, Average Merging with M$^3$ shows improvements of 7.31\% on AlpacaEval, 7.32\% on HumanEval, and 2.4\% on MBPP. 
Task Arithmetic with M$^3$ results in improvements of 2.0\% on AlpacaEval and 2.44\% on HumanEval for LM \& Code, and 10.4\% on MBPP for Math \& Code. 
TIES-Merging with M$^3$ achieves an improvement of 4.01\% for LM \& Math on GSM8K. For LM \& Code, TIES-Merging with M$^3$ shows significant improvements of 3.11\% on AlpacaEval, 25.61\% on HumanEval, and 30.8\% on MBPP.
Furthermore, as evidenced by the Avg. column in Table 
\ref{table:sft_models_merging_results}, M$^3$ consistently improves the overall performance of the merged models. This enhancement is particularly crucial for scenarios where the merged models must strike a balance between task-specific capabilities.
A few datasets do not show performance improvements with M$^3$, but the results remain on par with those without M$^3$, indicating that it does not degrade overall performance. Moreover, such cases are negligible compared to the overall benefits brought by M$^3$.

As shown in Table \ref{table:sft_models_merging_results}, M$^3$ achieves the best performance on four out of five task-specific datasets, surpassing even the individually fine-tuned LLMs (i.e., LM, Math, and Code). This suggests that M$^3$ enables the discovery of interpolated models along linear paths between two models that outperform both endpoints, providing support for the theoretical insights discussed in Section \ref{sec:Theoretical Analysis}.

In addition, we also compare our proposed method M$^3$ with existing model merging methods. The details are in Appendix~E.

\paragraph{Mitigating Suboptimal Fine-Tuning with M$^3$.}
\citet{yu2024language} noted that the poor performance of merging WizardMath-13B with llama-2-13b-code-alpaca stems from the latter not being sufficiently fine-tuned for code generation tasks. As shown in Table \ref{table:sft_models_merging_results}, the proposed M$^3$ method improves the pass@1 score on the code generation dataset MBPP by 10.4\%. This improvement demonstrates that even when one of the fine-tuned models is suboptimal for a specific task, M$^3$ can effectively unlock the merging potential by adaptively exploring suitable interpolation coefficients (i.e., contribution ratios), thereby maximizing the performance of the merged model. In other words, M$^3$ significantly mitigates the negative impact of insufficient fine-tuning on merging quality, further highlighting the critical importance of task-specific LLMs' contribution ratio in determining the performance of the merged model.

\begin{figure}[t]
  \centering  
  \includegraphics[width=1\columnwidth]{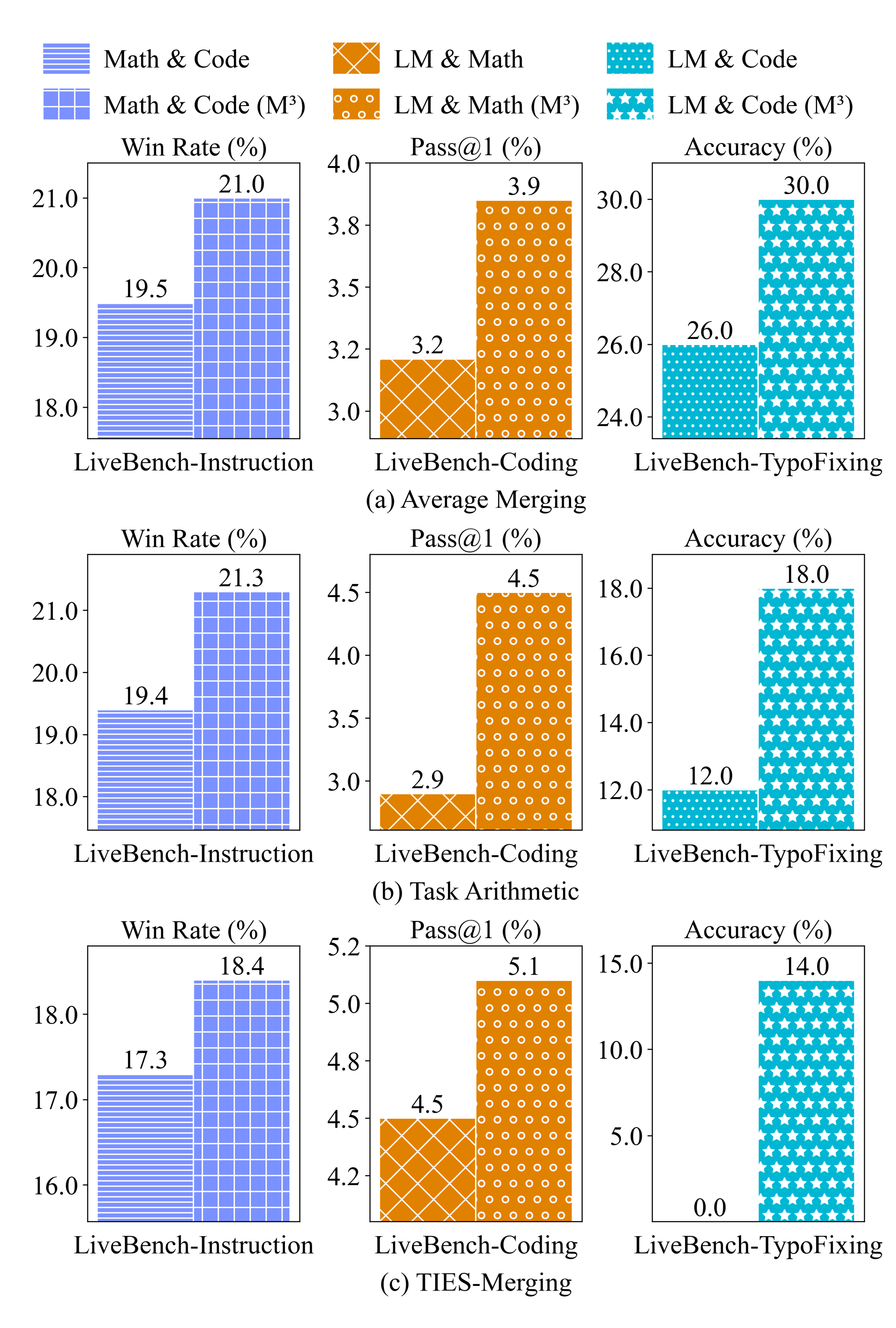}
  \caption{Performance of merged models (Math \& Code, LM \& Math, and LM \& Code) using three model merging methods (Average Merging, Task Arithmetic, and TIES-Merging) on OOD datasets.}
  \label{fig:ood_all} 
\end{figure}

\subsection{Model Robustness}
\paragraph{Out-of-Distribution Robustness.}
To ensure the evaluation datasets reflect true OOD scenarios, we select recently released datasets from domains not seen during fine-tuning. Accordingly, Math \& Code is evaluated on LiveBench-Instruction, LM \& Math on LiveBench-Coding, and LM \& Code on LiveBench-TypoFixing.
As shown in Figure~\ref{fig:ood_all}, M$^3$ consistently improves OOD performance across all model pairs and merging methods. For example, under Task Arithmetic, M$^3$ yields gains of 1.9\%, 1.6\%, and 6\% on the three respective tasks. Similar improvements are observed with Average Merging (1.5\%, 0.7\%, 4\%) and TIES-Merging (1.1\%, 0.6\%, 14\%). These results highlight M$^3$’s effectiveness in enhancing OOD generalization.    

\paragraph{Adversarial robustness.}
We evaluate the adversarial robustness of three merged models (Math \& Code, LM \& Math, and LM \& Code) using three prompt attack methods from PromptBench \citep{zhu2024promptbench}: DeepWordBug, BERTAttack, and StressTest. For DeepWordBug and BERTAttack, we randomly select three word positions per prompt. Robustness is measured by Performance Drop Rate (PDR) \citep{zhu2023promptrobust}, with lower PDR indicating stronger robustness (see Appendix~F for details).
As shown in Table~\ref{table:attack_results_stresstest}, M$^3$ notably improves robustness under StressTest attacks. For instance, it reduces PDR by 3.2\% (SST2) and 92.12\% (CoLA) for Math \& Code, and by 30.36\% (SST2) and 15.75\% (CoLA) for LM \& Code. Additionally, M$^3$ improves performance metrics such as accuracy and MCC—e.g., accuracy for LM \& Code on SST2 increases by 62.62\%.
These results indicate that M$^3$ enhances both robustness and task performance. Results for DeepWordBug and BERTAttack are provided in Appendix~G.

\subsection{Comparison of Mixup Model Merge and DARE}
DARE is a model sparsification method proposed by \citet{yu2024language} (see Appendix~H for details). We compare M$^3$ and DARE by integrating both into three representative model merging methods: Average Merging, Task Arithmetic, and TIES-Merging. The drop rate for DARE is fixed at 0.2.
As shown in Table~\ref{table:DARE_M3}, M$^3$ generally outperforms DARE and achieves significantly better results on several datasets. For example, on MBPP, the pass@1 score of the Math \& Code model increases from 9.8\% with DARE to 19\% with M$^3$.
Moreover, combining M$^3$ with DARE typically leads to even stronger performance, suggesting they are complementary. Notably, M$^3$ alone often yields the best results, while DARE alone rarely does, further highlighting the effectiveness of M$^3$.

\section{Conclusion}
Addressing the issue that existing methods overlook the varying contribution ratios of different models, we propose Mixup Model Merge (M$^3$), a novel approach based on the randomized linear interpolation strategy of Mixup. M$^3$ performs linear interpolation in the parameter space between two task-specific large language models, with interpolation coefficients sampled from a Beta distribution to explore a diverse range of contribution ratios. This method leverages controllable randomness to significantly outperform traditional equal-ratio merging. Extensive experiments demonstrate that M$^3$ not only improves merged model performance across multiple tasks but also enhances out-of-distribution and adversarial robustness, and effectively combines with model sparsification methods such as DARE. By tuning the Beta distribution parameters, M$^3$ balances efficiency and diversity in exploring contribution ratios. This simple and efficient method offers new insights for optimizing model merging strategies.

In the future, M$^3$ may also be extended to broader applications, such as merging fine-tuned models with Reinforcement Learning with Human Feedback (RLHF) models~\citep{ouyang2022training}, potentially reducing the alignment tax associated with post-hoc alignment techniques like fine-tuning.

\bibliography{aaai2026}

\begin{thebibliography}{50}
\providecommand{\natexlab}[1]{#1}

\bibitem[{Akiba et~al.(2024)Akiba, Shing, Tang, Sun, and Ha}]{akiba2024evolutionary}
Akiba, T.; Shing, M.; Tang, Y.; Sun, Q.; and Ha, D. 2024.
\newblock Evolutionary optimization of model merging recipes.
\newblock \emph{arXiv preprint arXiv:2403.13187}.

\bibitem[{Austin et~al.(2021)Austin, Odena, Nye, Bosma, Michalewski, Dohan, Jiang, Cai, Terry, Le et~al.}]{austin2021program}
Austin, J.; Odena, A.; Nye, M.; Bosma, M.; Michalewski, H.; Dohan, D.; Jiang, E.; Cai, C.; Terry, M.; Le, Q.; et~al. 2021.
\newblock Program synthesis with large language models.
\newblock \emph{arXiv preprint arXiv:2108.07732}.

\bibitem[{Beeching et~al.(2023)Beeching, Fourrier, Habib, Han, Lambert, Rajani, Sanseviero, Tunstall, and Wolf}]{beeching2023open}
Beeching, E.; Fourrier, C.; Habib, N.; Han, S.; Lambert, N.; Rajani, N.; Sanseviero, O.; Tunstall, L.; and Wolf, T. 2023.
\newblock Open llm leaderboard.
\newblock \emph{Hugging Face}.

\bibitem[{Brown et~al.(2020)Brown, Mann, Ryder, Subbiah, Kaplan, Dhariwal, Neelakantan, Shyam, Sastry, Askell et~al.}]{brown2020language}
Brown, T.; Mann, B.; Ryder, N.; Subbiah, M.; Kaplan, J.~D.; Dhariwal, P.; Neelakantan, A.; Shyam, P.; Sastry, G.; Askell, A.; et~al. 2020.
\newblock Language models are few-shot learners.
\newblock \emph{Advances in neural information processing systems}, 33: 1877--1901.

\bibitem[{Chang, Chang, and Wu(2024)}]{chang2024ba}
Chang, Y.; Chang, Y.; and Wu, Y. 2024.
\newblock BA-LoRA: Bias-Alleviating Low-Rank Adaptation to Mitigate Catastrophic Inheritance in Large Language Models.
\newblock \emph{arXiv preprint arXiv:2408.04556}.

\bibitem[{Chang et~al.(2024)Chang, Wang, Wang, Wu, Yang, Zhu, Chen, Yi, Wang, Wang et~al.}]{chang2024survey}
Chang, Y.; Wang, X.; Wang, J.; Wu, Y.; Yang, L.; Zhu, K.; Chen, H.; Yi, X.; Wang, C.; Wang, Y.; et~al. 2024.
\newblock A survey on evaluation of large language models.
\newblock \emph{ACM Transactions on Intelligent Systems and Technology}, 15(3): 1--45.

\bibitem[{Chaudhary(2023)}]{chaudhary2023code}
Chaudhary, S. 2023.
\newblock Code alpaca: An instruction-following llama model for code generation.
\newblock \emph{GitHub repository}.

\bibitem[{Chen et~al.(2021)Chen, Tworek, Jun, Yuan, Pinto, Kaplan, Edwards, Burda, Joseph, Brockman et~al.}]{chen2021evaluating}
Chen, M.; Tworek, J.; Jun, H.; Yuan, Q.; Pinto, H. P. D.~O.; Kaplan, J.; Edwards, H.; Burda, Y.; Joseph, N.; Brockman, G.; et~al. 2021.
\newblock Evaluating large language models trained on code.
\newblock \emph{arXiv preprint arXiv:2107.03374}.

\bibitem[{Chowdhery et~al.(2023)Chowdhery, Narang, Devlin, Bosma, Mishra, Roberts, Barham, Chung, Sutton, Gehrmann et~al.}]{chowdhery2023palm}
Chowdhery, A.; Narang, S.; Devlin, J.; Bosma, M.; Mishra, G.; Roberts, A.; Barham, P.; Chung, H.~W.; Sutton, C.; Gehrmann, S.; et~al. 2023.
\newblock Palm: Scaling language modeling with pathways.
\newblock \emph{Journal of Machine Learning Research}, 24(240): 1--113.

\bibitem[{Cobbe et~al.(2021)Cobbe, Kosaraju, Bavarian, Chen, Jun, Kaiser, Plappert, Tworek, Hilton, Nakano et~al.}]{cobbe2021training}
Cobbe, K.; Kosaraju, V.; Bavarian, M.; Chen, M.; Jun, H.; Kaiser, L.; Plappert, M.; Tworek, J.; Hilton, J.; Nakano, R.; et~al. 2021.
\newblock Training verifiers to solve math word problems.
\newblock \emph{arXiv preprint arXiv:2110.14168}.

\bibitem[{Deep, Bhardwaj, and Poria(2024)}]{deep2024della}
Deep, P.~T.; Bhardwaj, R.; and Poria, S. 2024.
\newblock Della-merging: Reducing interference in model merging through magnitude-based sampling.
\newblock \emph{arXiv preprint arXiv:2406.11617}.

\bibitem[{Ding et~al.(2023)Ding, Qin, Yang, Wei, Yang, Su, Hu, Chen, Chan, Chen et~al.}]{ding2023parameter}
Ding, N.; Qin, Y.; Yang, G.; Wei, F.; Yang, Z.; Su, Y.; Hu, S.; Chen, Y.; Chan, C.-M.; Chen, W.; et~al. 2023.
\newblock Parameter-efficient fine-tuning of large-scale pre-trained language models.
\newblock \emph{Nature Machine Intelligence}, 5(3): 220--235.

\bibitem[{Dubois et~al.(2024)Dubois, Galambosi, Liang, and Hashimoto}]{dubois2024length}
Dubois, Y.; Galambosi, B.; Liang, P.; and Hashimoto, T.~B. 2024.
\newblock Length-controlled alpacaeval: A simple way to debias automatic evaluators.
\newblock \emph{arXiv preprint arXiv:2404.04475}.

\bibitem[{Fisher(1922)}]{fisher1922mathematical}
Fisher, R.~A. 1922.
\newblock On the mathematical foundations of theoretical statistics.
\newblock \emph{Philosophical transactions of the Royal Society of London. Series A, containing papers of a mathematical or physical character}, 222(594-604): 309--368.

\bibitem[{Gao et~al.(2018)Gao, Lanchantin, Soffa, and Qi}]{gao2018black}
Gao, J.; Lanchantin, J.; Soffa, M.~L.; and Qi, Y. 2018.
\newblock Black-box generation of adversarial text sequences to evade deep learning classifiers.
\newblock In \emph{2018 IEEE Security and Privacy Workshops (SPW)}, 50--56. IEEE.

\bibitem[{Guo et~al.(2024)Guo, Xu, Chang, and Wu}]{guo2024chbench}
Guo, C.; Xu, N.; Chang, Y.; and Wu, Y. 2024.
\newblock Chbench: A chinese dataset for evaluating health in large language models.
\newblock \emph{arXiv preprint arXiv:2409.15766}.

\bibitem[{Hendrycks et~al.(2021)Hendrycks, Burns, Kadavath, Arora, Basart, Tang, Song, and Steinhardt}]{hendrycks2021measuring}
Hendrycks, D.; Burns, C.; Kadavath, S.; Arora, A.; Basart, S.; Tang, E.; Song, D.; and Steinhardt, J. 2021.
\newblock Measuring mathematical problem solving with the math dataset.
\newblock \emph{arXiv preprint arXiv:2103.03874}.

\bibitem[{Hu et~al.(2021)Hu, Shen, Wallis, Allen-Zhu, Li, Wang, Wang, and Chen}]{hu2021lora}
Hu, E.~J.; Shen, Y.; Wallis, P.; Allen-Zhu, Z.; Li, Y.; Wang, S.; Wang, L.; and Chen, W. 2021.
\newblock Lora: Low-rank adaptation of large language models.
\newblock \emph{arXiv preprint arXiv:2106.09685}.

\bibitem[{Ilharco et~al.(2022)Ilharco, Ribeiro, Wortsman, Gururangan, Schmidt, Hajishirzi, and Farhadi}]{ilharco2022editing}
Ilharco, G.; Ribeiro, M.~T.; Wortsman, M.; Gururangan, S.; Schmidt, L.; Hajishirzi, H.; and Farhadi, A. 2022.
\newblock Editing models with task arithmetic.
\newblock \emph{arXiv preprint arXiv:2212.04089}.

\bibitem[{Jiao et~al.(2023)Jiao, Wang, Huang, Wang, and Tu}]{jiao2023chatgpt}
Jiao, W.; Wang, W.; Huang, J.-t.; Wang, X.; and Tu, Z. 2023.
\newblock Is ChatGPT a good translator? A preliminary study.
\newblock \emph{arXiv preprint arXiv:2301.08745}, 1(10).

\bibitem[{Jin et~al.(2022)Jin, Ren, Preotiuc-Pietro, and Cheng}]{jin2022dataless}
Jin, X.; Ren, X.; Preotiuc-Pietro, D.; and Cheng, P. 2022.
\newblock Dataless knowledge fusion by merging weights of language models.
\newblock \emph{arXiv preprint arXiv:2212.09849}.

\bibitem[{Li et~al.(2020)Li, Ma, Guo, Xue, and Qiu}]{li2020bert}
Li, L.; Ma, R.; Guo, Q.; Xue, X.; and Qiu, X. 2020.
\newblock Bert-attack: Adversarial attack against bert using bert.
\newblock \emph{arXiv preprint arXiv:2004.09984}.

\bibitem[{Li et~al.(2023)Li, Zhang, Dubois, Taori, Gulrajani, Guestrin, Liang, and Hashimoto}]{li2023alpacaeval}
Li, X.; Zhang, T.; Dubois, Y.; Taori, R.; Gulrajani, I.; Guestrin, C.; Liang, P.; and Hashimoto, T.~B. 2023.
\newblock Alpacaeval: An automatic evaluator of instruction-following models.

\bibitem[{Lin et~al.(2023)Lin, Lin, Xiong, Diao, Liu, Zhang, Pan, Wang, Hu, Zhang et~al.}]{lin2023mitigating}
Lin, Y.; Lin, H.; Xiong, W.; Diao, S.; Liu, J.; Zhang, J.; Pan, R.; Wang, H.; Hu, W.; Zhang, H.; et~al. 2023.
\newblock Mitigating the alignment tax of rlhf.
\newblock \emph{arXiv preprint arXiv:2309.06256}.

\bibitem[{Lin et~al.(2024)Lin, Lin, Xiong, Diao, Liu, Zhang, Pan, Wang, Hu, Zhang et~al.}]{lin2024mitigating}
Lin, Y.; Lin, H.; Xiong, W.; Diao, S.; Liu, J.; Zhang, J.; Pan, R.; Wang, H.; Hu, W.; Zhang, H.; et~al. 2024.
\newblock Mitigating the alignment tax of rlhf.
\newblock In \emph{Proceedings of the 2024 Conference on Empirical Methods in Natural Language Processing}, 580--606.

\bibitem[{Luo et~al.(2023)Luo, Sun, Xu, Zhao, Lou, Tao, Geng, Lin, Chen, and Zhang}]{luo2023wizardmath}
Luo, H.; Sun, Q.; Xu, C.; Zhao, P.; Lou, J.; Tao, C.; Geng, X.; Lin, Q.; Chen, S.; and Zhang, D. 2023.
\newblock Wizardmath: Empowering mathematical reasoning for large language models via reinforced evol-instruct.
\newblock \emph{arXiv preprint arXiv:2308.09583}.

\bibitem[{Matena and Raffel(2022)}]{matena2022merging}
Matena, M.~S.; and Raffel, C.~A. 2022.
\newblock Merging models with fisher-weighted averaging.
\newblock \emph{Advances in Neural Information Processing Systems}, 35: 17703--17716.

\bibitem[{Naik et~al.(2018)Naik, Ravichander, Sadeh, Rose, and Neubig}]{naik2018stress}
Naik, A.; Ravichander, A.; Sadeh, N.; Rose, C.; and Neubig, G. 2018.
\newblock Stress test evaluation for natural language inference.
\newblock \emph{arXiv preprint arXiv:1806.00692}.

\bibitem[{Nam et~al.(2024)Nam, Macvean, Hellendoorn, Vasilescu, and Myers}]{nam2024using}
Nam, D.; Macvean, A.; Hellendoorn, V.; Vasilescu, B.; and Myers, B. 2024.
\newblock Using an llm to help with code understanding.
\newblock In \emph{Proceedings of the IEEE/ACM 46th International Conference on Software Engineering}, 1--13.

\bibitem[{Neyshabur, Sedghi, and Zhang(2020)}]{neyshabur2020being}
Neyshabur, B.; Sedghi, H.; and Zhang, C. 2020.
\newblock What is being transferred in transfer learning?
\newblock \emph{Advances in neural information processing systems}, 33: 512--523.

\bibitem[{OpenAI(2023)}]{openai2023gpt}
OpenAI, R. 2023.
\newblock Gpt-4 technical report. arxiv 2303.08774.
\newblock \emph{View in Article}, 2(5).

\bibitem[{Ouyang et~al.(2022)Ouyang, Wu, Jiang, Almeida, Wainwright, Mishkin, Zhang, Agarwal, Slama, Ray et~al.}]{ouyang2022training}
Ouyang, L.; Wu, J.; Jiang, X.; Almeida, D.; Wainwright, C.; Mishkin, P.; Zhang, C.; Agarwal, S.; Slama, K.; Ray, A.; et~al. 2022.
\newblock Training language models to follow instructions with human feedback.
\newblock \emph{Advances in neural information processing systems}, 35: 27730--27744.

\bibitem[{Socher et~al.(2013)Socher, Perelygin, Wu, Chuang, Manning, Ng, and Potts}]{socher2013recursive}
Socher, R.; Perelygin, A.; Wu, J.; Chuang, J.; Manning, C.~D.; Ng, A.~Y.; and Potts, C. 2013.
\newblock Recursive deep models for semantic compositionality over a sentiment treebank.
\newblock In \emph{Proceedings of the 2013 conference on empirical methods in natural language processing}, 1631--1642.

\bibitem[{Taori et~al.(2023)Taori, Gulrajani, Zhang, Dubois, Li, Guestrin, Liang, and Hashimoto}]{alpaca}
Taori, R.; Gulrajani, I.; Zhang, T.; Dubois, Y.; Li, X.; Guestrin, C.; Liang, P.; and Hashimoto, T.~B. 2023.
\newblock Stanford Alpaca: An Instruction-following LLaMA model.
\newblock \url{https://github.com/tatsu-lab/stanford_alpaca}.

\bibitem[{Touvron et~al.(2023)Touvron, Martin, Stone, Albert, Almahairi, Babaei, Bashlykov, Batra, Bhargava, Bhosale et~al.}]{touvron2023llama}
Touvron, H.; Martin, L.; Stone, K.; Albert, P.; Almahairi, A.; Babaei, Y.; Bashlykov, N.; Batra, S.; Bhargava, P.; Bhosale, S.; et~al. 2023.
\newblock Llama 2: Open foundation and fine-tuned chat models.
\newblock \emph{arXiv preprint arXiv:2307.09288}.

\bibitem[{Wang et~al.(2023)Wang, Hu, Hou, Chen, Zheng, Wang, Yang, Huang, Ye, Geng et~al.}]{wang2023robustness}
Wang, J.; Hu, X.; Hou, W.; Chen, H.; Zheng, R.; Wang, Y.; Yang, L.; Huang, H.; Ye, W.; Geng, X.; et~al. 2023.
\newblock On the robustness of chatgpt: An adversarial and out-of-distribution perspective.
\newblock \emph{arXiv preprint arXiv:2302.12095}.

\bibitem[{Wang et~al.(2022)Wang, Kordi, Mishra, Liu, Smith, Khashabi, and Hajishirzi}]{wang2022self}
Wang, Y.; Kordi, Y.; Mishra, S.; Liu, A.; Smith, N.~A.; Khashabi, D.; and Hajishirzi, H. 2022.
\newblock Self-instruct: Aligning language models with self-generated instructions.
\newblock \emph{arXiv preprint arXiv:2212.10560}.

\bibitem[{Warstadt(2019)}]{warstadt2019neural}
Warstadt, A. 2019.
\newblock Neural Network Acceptability Judgments.
\newblock \emph{arXiv preprint arXiv:1805.12471}.

\bibitem[{Wei et~al.(2022)Wei, Wang, Schuurmans, Bosma, Xia, Chi, Le, Zhou et~al.}]{wei2022chain}
Wei, J.; Wang, X.; Schuurmans, D.; Bosma, M.; Xia, F.; Chi, E.; Le, Q.~V.; Zhou, D.; et~al. 2022.
\newblock Chain-of-thought prompting elicits reasoning in large language models.
\newblock \emph{Advances in neural information processing systems}, 35: 24824--24837.

\bibitem[{White et~al.(2024)White, Dooley, Roberts, Pal, Feuer, Jain, Shwartz-Ziv, Jain, Saifullah, Naidu et~al.}]{white2024livebench}
White, C.; Dooley, S.; Roberts, M.; Pal, A.; Feuer, B.; Jain, S.; Shwartz-Ziv, R.; Jain, N.; Saifullah, K.; Naidu, S.; et~al. 2024.
\newblock Livebench: A challenging, contamination-free llm benchmark.
\newblock \emph{arXiv preprint arXiv:2406.19314}.

\bibitem[{Wortsman et~al.(2022)Wortsman, Ilharco, Gadre, Roelofs, Gontijo-Lopes, Morcos, Namkoong, Farhadi, Carmon, Kornblith et~al.}]{wortsman2022model}
Wortsman, M.; Ilharco, G.; Gadre, S.~Y.; Roelofs, R.; Gontijo-Lopes, R.; Morcos, A.~S.; Namkoong, H.; Farhadi, A.; Carmon, Y.; Kornblith, S.; et~al. 2022.
\newblock Model soups: averaging weights of multiple fine-tuned models improves accuracy without increasing inference time.
\newblock In \emph{International conference on machine learning}, 23965--23998. PMLR.

\bibitem[{Xia et~al.(2024)Xia, Yu, Dang, Yang, Wu, Tian, Chang, and Lin}]{xia2024rethinking}
Xia, T.; Yu, B.; Dang, K.; Yang, A.; Wu, Y.; Tian, Y.; Chang, Y.; and Lin, J. 2024.
\newblock Rethinking data selection at scale: Random selection is almost all you need.
\newblock \emph{arXiv preprint arXiv:2410.09335}.

\bibitem[{Xing(2024)}]{xing2024designing}
Xing, F. 2024.
\newblock Designing heterogeneous llm agents for financial sentiment analysis.
\newblock \emph{ACM Transactions on Management Information Systems}.

\bibitem[{Xu et~al.(2024)Xu, Sun, Zheng, Geng, Zhao, Feng, Tao, Lin, and Jiang}]{xu2024wizardlm}
Xu, C.; Sun, Q.; Zheng, K.; Geng, X.; Zhao, P.; Feng, J.; Tao, C.; Lin, Q.; and Jiang, D. 2024.
\newblock WizardLM: Empowering large pre-trained language models to follow complex instructions.
\newblock In \emph{The Twelfth International Conference on Learning Representations}.

\bibitem[{Yadav et~al.(2023)Yadav, Tam, Choshen, Raffel, and Bansal}]{yadav2023resolving}
Yadav, P.; Tam, D.; Choshen, L.; Raffel, C.; and Bansal, M. 2023.
\newblock Resolving interference when merging models.
\newblock \emph{arXiv preprint arXiv:2306.01708}, 1.

\bibitem[{Yang et~al.(2024)Yang, Shen, Guo, Wang, Cao, Zhang, and Tao}]{yang2024model}
Yang, E.; Shen, L.; Guo, G.; Wang, X.; Cao, X.; Zhang, J.; and Tao, D. 2024.
\newblock Model merging in llms, mllms, and beyond: Methods, theories, applications and opportunities.
\newblock \emph{arXiv preprint arXiv:2408.07666}.

\bibitem[{Yu et~al.(2024)Yu, Yu, Yu, Huang, and Li}]{yu2024language}
Yu, L.; Yu, B.; Yu, H.; Huang, F.; and Li, Y. 2024.
\newblock Language models are super mario: Absorbing abilities from homologous models as a free lunch.
\newblock In \emph{Forty-first International Conference on Machine Learning}.

\bibitem[{Zhang(2017)}]{zhang2017mixup}
Zhang, H. 2017.
\newblock mixup: Beyond empirical risk minimization.
\newblock \emph{arXiv preprint arXiv:1710.09412}.

\bibitem[{Zhu et~al.(2023)Zhu, Wang, Zhou, Wang, Chen, Wang, Yang, Ye, Zhang, Gong et~al.}]{zhu2023promptrobust}
Zhu, K.; Wang, J.; Zhou, J.; Wang, Z.; Chen, H.; Wang, Y.; Yang, L.; Ye, W.; Zhang, Y.; Gong, N.; et~al. 2023.
\newblock Promptrobust: Towards evaluating the robustness of large language models on adversarial prompts.
\newblock In \emph{Proceedings of the 1st ACM Workshop on Large AI Systems and Models with Privacy and Safety Analysis}, 57--68.

\bibitem[{Zhu et~al.(2024)Zhu, Zhao, Chen, Wang, and Xie}]{zhu2024promptbench}
Zhu, K.; Zhao, Q.; Chen, H.; Wang, J.; and Xie, X. 2024.
\newblock Promptbench: A unified library for evaluation of large language models.
\newblock \emph{Journal of Machine Learning Research}, 25(254): 1--22.

\end{thebibliography}

\clearpage
\appendix

\section{Task-Specific Fine-Tuned LLMs and Datasets Details}
\label{sec:Task-Specific Fine-Tuned LLMs and Datasets Details}
We conduct model merging experiments using three task-specific LLMs fine-tuned from Llama-2-13b:
\begin{itemize}
\item \textbf{WizardLM-13B} is an instruction-following model based on Llama-2-13b, designed to improve open-domain instruction-following. Using the Evol-Instruct method \citep{xu2024wizardlm}, it generates high-complexity instruction data to reduce human annotation and enhance generalization. The model undergoes supervised fine-tuning with AI-generated data, followed by refinement via RLHF. Evaluation results show that Evol-Instruct-generated instructions outperform human-written ones, and WizardLM-13B surpasses ChatGPT in high-complexity tasks. In GPT-4 automated evaluation, it achieves over 90\% of ChatGPT's performance in 17 out of 29 tasks, demonstrating the effectiveness of AI-evolved instruction fine-tuning \citep{xu2024wizardlm}.

\item \textbf{WizardMath-13B}, optimized from Llama-2-13b, is designed for mathematical reasoning and enhances Chain-of-Thought (CoT) \citep{wei2022chain} capabilities. It uses Reinforcement Learning from Evol-Instruct Feedback to evolve math tasks and improve reasoning. Trained on GSM8K and MATH datasets, it excels in both basic and advanced math problems. In evaluations, WizardMath-Mistral 7B outperforms all open-source models with fewer training data, while WizardMath 70B surpasses GPT-3.5-Turbo, Claude 2, and even early GPT-4 versions in mathematical reasoning tasks.

\item \textbf{llama-2-13b-code-alpaca} is a code generation model fine-tuned from Llama-2-13b, designed to enhance code understanding and generation. It follows the same training approach as Stanford Alpaca \citep{alpaca} but focuses on code-related tasks. The model is fine-tuned with 20K instruction-following code samples generated using the Self-Instruct method \citep{wang2022self}. However, as it has not undergone safety fine-tuning, caution is required when using it in production environments.
\end{itemize}

We use one dataset to evaluate the instruction-following task:
\begin{itemize}
\item \textbf{AlpacaEval} \citep{li2023alpacaeval} is an LLM-based automated evaluation metric that assesses model performance by testing on a fixed set of 805 instructions and computing the win rate of the evaluated model against a baseline. The evaluation process involves an LLM-based evaluator that compares the responses and determines the probability of preferring the evaluated model.
In this paper, we use AlpacaEval 2.0 \citep{dubois2024length}. To reduce costs, we use chatgpt\_fn for evaluation.
\end{itemize}

We use two dataset to evaluate the mathematical reasoning task:
\begin{itemize}
\item \textbf{GSM8K} is a dataset of 8.5K high-quality, linguistically diverse grade school math word problems, designed to evaluate the multi-step mathematical reasoning abilities of large language models. It consists of 7.5K training problems and 1K test problems. In this paper, we use the 1K test set for evaluation \citep{cobbe2021training}.

\item \textbf{MATH} is a dataset containing 12,500 competition-level math problems, designed to evaluate and enhance the problem-solving abilities of machine learning models. It consists of 7,500 training problems and 5,000 test problems. We use the 5,000 test set for evaluation \citep{hendrycks2021measuring}. 
\end{itemize}

We used two dataset to evaluate the code generation task:
\begin{itemize}
\item \textbf{HumanEval} is a dataset consisting of 164 hand-written programming problems, designed to evaluate the functional correctness of code generation models. Each problem includes a function signature, docstring, function body, and unit tests. The dataset tests models' language comprehension, reasoning, and algorithmic abilities \citep{chen2021evaluating}.

\item \textbf{MBPP} is a dataset containing 974 programming problems designed to evaluate a model's ability to synthesize Python programs from natural language descriptions. The problems range from basic numerical operations to more complex tasks involving list and string processing. The test set consists of 500 problems, which are used for evaluation in this paper \citep{austin2021program}.
\end{itemize}

\section{Out-of-Distribution Dataset Selection Details}
\label{sec:Out-of-Distribution Dataset Selection Details}
LiveBench \citep{white2024livebench} is a dynamic benchmark for large language models, featuring frequently updated questions and diverse tasks. To assess OOD robustness, we evaluate math \& code, LM \& math, and LM \& code models using instruction following (LiveBench-Instruction), coding (LiveBench-Coding), and language comprehension (LiveBench-TypoFixing) category in LiveBench, respectively, deliberately avoiding the fine-tuning domains of the merged fine-tuned models.
These tasks were released after November 2023, whereas WizardLM-13B, WizardMath-13B, and llama-2-13b-code-alpaca were all introduced earlier. Furthermore, their shared Llama-2-13b backbone was trained on data only up to July 2023. Consequently, these factors collectively ensure that the evaluation remains OOD in the temporal dimension.

When assessing the OOD robustness of LM \& Code using the Language Comprehension category in LiveBench, only the typo-fixing task is considered. This decision is based on the fact that LiveBench is highly challenging, and the merged model performs poorly on other tasks in this category, with accuracy close to zero, rendering the evaluation results inconclusive and uninformative.

Finally, we acknowledge the limitations of these datasets. For large models like Llama-2-13b, identifying truly OOD datasets is difficult, as their training data likely covers similar distributions. These datasets are better described as "out-of-example", representing instances not explicitly seen during training. As discussed in \citep{wang2023robustness}, distribution shifts can occur across domains and time. While Llama-2-13b may have been trained on datasets for tasks like instruction-following, coding, and language comprehension, the datasets we selected remain valuable for OOD evaluation by capturing temporal shifts, providing insights into robustness over time.

\section{Adversarial Robustness Evaluation Experiments Setting Details}
\label{sec:Adversarial Robustness Evaluation Experiments Setting Details}
PromptBench \citep{zhu2024promptbench} is a unified library designed for evaluating LLMs, providing a standardized and extensible framework. It includes several key components such as prompt construction, prompt engineering, dataset and model loading, adversarial prompt attacks, dynamic evaluation protocols, and analysis tools. 

We use the Adversarial Prompt Attacks module in PromptBench aims to evaluate the robustness of LLMs against adversarial prompts. We employ three methods to perform adversarial attacks on prompts to evaluate the adversarial robustness of the merged models: DeepWordBug \citep{gao2018black}, BERTAttack \citep{li2020bert}, and StressTest \citep{naik2018stress}, representing Character-level, Word-level, and Sentence-level attacks, respectively.
\begin{itemize}
\item \textbf{DeepWordBug} introduces subtle character-level perturbations (e.g., adding, deleting, or replacing characters) to words in text to deceive language models. It aims to evaluate a model's robustness against small typographical errors that may alter the model's performance without being easily detected.
\item \textbf{BERTAttack} manipulates text at the word level by replacing words with contextually similar synonyms to mislead large language models. This method tests the model's ability to maintain accuracy despite small lexical changes that might alter the meaning of the input.
\item \textbf{StressTest} appends irrelevant or redundant sentences to the end of a prompt to distract and confuse language models. It assesses the model’s ability to handle extraneous information and maintain accuracy when faced with unnecessary distractions.
\end{itemize}

\begin{figure}[t]
  \centering
  \begin{subfigure}[b]{\columnwidth}
    \centering
    \includegraphics[width=\columnwidth]{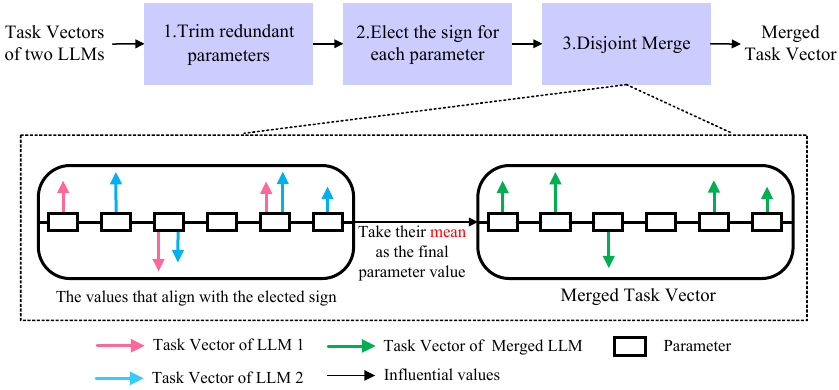}
    \captionsetup{skip=2pt}  
    \caption{The operational steps of TIES-Merging.}
    \label{fig:TIES-Merging}
  \end{subfigure}

  \vspace{0.3cm} 

  \begin{subfigure}[b]{\columnwidth}
    \centering
    \includegraphics[width=\columnwidth]{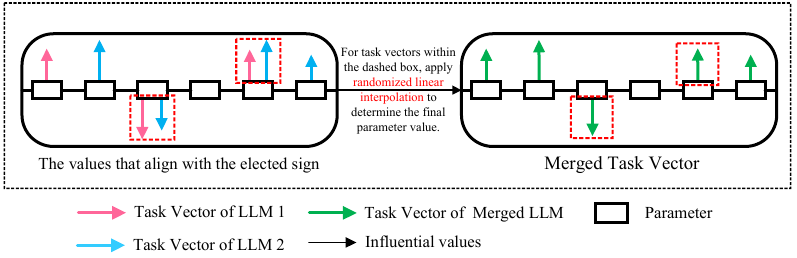}
    \captionsetup{skip=2pt}  
    \caption{After introducing M$^3$, the Disjoint Merge step in the TIES-Merging procedure.}
    \label{fig:TIES-Merging+M$^3$}
  \end{subfigure}
 
  \vspace{0.1cm} 
  
  \caption{The difference between M$^3$ and the original TIES-Merging is that, in the Disjoint Merge step, when two task vectors are retained for a given parameter, the mean of the task vectors is replaced by a random linear interpolation, while the other operations remain unchanged.}
  \label{fig:ties}
\end{figure}
The evaluation is conducted on the Sentiment Analysis dataset (SST2 \citep{socher2013recursive}) and the Grammar Correctness dataset (CoLA \citep{warstadt2019neural}):
\begin{itemize}
\item \textbf{SST2} \citep{socher2013recursive}: A sentiment analysis dataset designed to assess whether a given sentence conveys a positive or negative sentiment.
\item \textbf{CoLA} \citep{warstadt2019neural}: A dataset for grammar correctness, where the model must determine whether a sentence is grammatically acceptable.
\end{itemize}

\section{Hyperparameter Setting Details in Model Merging Methods}
\label{sec:Hyperparameter Setting Details in Model Merging Methods}
Table~\ref{table:hyperparameter_search} presents the hyperparameter search ranges for the model merging methods. For Task Arithmetic and TIES-Merging, the scaling terms are selected from $[0.5, 1.0]$, while in TIES-Merging, the retain ratio for the largest-magnitude parameters is chosen from $[0.5, 0.7, 0.9]$. In contrast, the Average Merging method does not require any hyperparameters.
\begin{table}[t]
\small 
\centering
\resizebox{\columnwidth}{!}{
\begin{tabular}{cc}  
\toprule
Merging Methods & Search Ranges of Hyperparameters \\ 
\midrule
\multirow{2}{*}{Task Arithmetic} & Scaling term for merging model parameters: \\
                                 & [0.5, 0.6, 0.7, 0.8, 0.9, 1.0] \\ \midrule
\multirow{4}{*}{TIES-Merging} & Scaling term for merging model parameters: \\
                              & [0.5, 0.6, 0.7, 0.8, 0.9, 1.0] \\  \cline{2-2}
                              & Ratio for retaining parameters with the  \\
                              & largest-magnitude values: [0.5, 0.7, 0.9] \\  
\bottomrule
\end{tabular}}
\caption{\label{table:hyperparameter_search} Hyperparameter search ranges for model merging methods.}
\end{table}

Table~\ref{table:optimal hyperparameter} presents the optimal hyperparameter settings for the TIES-Merging model merging method obtained through searching. These settings are further applied to model merging experiments involving M$^3$ and DARE.
\begin{table}[t]
\centering
\resizebox{\columnwidth}{!}{
\begin{tabular}{ccc} 
\toprule
\makecell{Merging Method} & Model & \makecell{Hyperparameter Values} \\ 
\midrule
\multirow{3}{*}{\makecell{TIES-Merging}} 
    & \makecell{LM \& Math} & \makecell{scaling\_term=0.5, retain\_ratio=0.5} \\ 
\cline{2-3}
    & \makecell{LM \& Code} & \makecell{scaling\_term=1.0, retain\_ratio=0.7} \\ 
\cline{2-3}
    & \makecell{Math \& Code} & \makecell{scaling\_term=1.0, retain\_ratio=0.5} \\ 
\bottomrule
\end{tabular}}
\caption{\label{table:optimal hyperparameter} Hyperparameter settings in TIES-Merging.}
\end{table}

\section{Comparison of M$^3$ and Existing Model Merging Approaches}
\label{sec:compare_to_baseline}
\begin{table*}[t]
\centering
\small 
\resizebox{0.8\textwidth}{!}{
\begin{tabular}{c c c c c c c c} \toprule

\multirow{2}{*}{\makecell{Models}} &
\multirow{2}{*}{\makecell{Merging Methods}} & 
\makecell{Instruction\\ Following} & 
\makecell{Mathematical \\ Reasoning} &
\makecell{Code\\ Generating} &
\multirow{2}{*}{\makecell{Avg.}} \\ \cmidrule(lr){3-3} \cmidrule(lr){4-4} \cmidrule(lr){5-5}

& & AlpacaEval & GSM8K & MBPP & \\ \hline
                   
\multirow{6}{*}{\makecell{LM \\ \& Math}} & \makecell{Average Merging} & 45.28 & 66.34& - & 55.81 \\ \cline{2-6}

                                           & \makecell{Task Arithmetic} & \underline{45.78} & 66.34& - & 56.06 \\ \cline{2-6}

                                           & \makecell{TIES-Merging} & 38.63& 14.56& - & 26.60 \\  \cline{2-6}

                                           & DARE & \textbf{49.00}& \underline{66.64}& - & \textbf{57.82} \\ \cline{2-6}

                                           & DELLA & 44.03 & 48.45 & - & 46.24 \\ \cline{2-6}

                                           & M$^3$ (Ours) & 44.90 & \textbf{67.32} & - & \underline{56.11} \\ \cline{1-6}

\multirow{6}{*}{\makecell{LM \\ \& Code}} & \makecell{Average Merging} & 36.60& - & 32.00 & 34.3 \\ \cline{2-6}

                                           & \makecell{Task Arithmetic} & \underline{44.64}& - & 33.60 & \underline{39.12} \\ \cline{2-6}

                                           & \makecell{TIES-Merging} & 41.85& - & 0.0 & 20.93 \\  \cline{2-6}

                                           & DARE & 41.47 & - & 33.00 & 37.24 \\ \cline{2-6}

                                           & DELLA & 40.99 & - & \underline{35.00} & 38.00 \\ \cline{2-6}

                                           & M$^3$ (Ours) & \textbf{45.20} & - & \textbf{35.20} & \textbf{40.2} \\ \cline{1-6}

\multirow{6}{*}{\makecell{Math \\ \& Code}} & \makecell{Average Merging} & - & 56.17& 8.20 & 32.19 \\ \cline{2-6}

                                           & \makecell{Task Arithmetic} & - & 64.67& 8.60 & 36.64 \\ \cline{2-6}

                                           & \makecell{TIES-Merging} & - & 64.67& \textbf{22.60} & \textbf{43.64} \\  \cline{2-6}

                                           & DARE & - & \textbf{65.05} & 9.80 & 37.43 \\ \cline{2-6}

                                           & DELLA & - & 63.08 & 20.2 & 41.64 \\  \cline{2-6}
                                            
                                           & M$^3$ (Ours) & - & \underline{64.75} & \underline{21.4} & \underline{43.10} \\ \bottomrule
\end{tabular}}
\caption{\label{table:compare_to_baseline} Comparison of our method M$^3$ and existing model merging approaches. The best and second-best results are marked in bold and underlined fonts.}
\end{table*}

For each baseline method—Average Merging, Task Arithmetic, TIES-Merging, and DARE—we combine it with our proposed M$^3$ method and select the best-performing combination for each merged model (LM \& Math, LM \& Code and Math \& Code). These optimal results are reported as M$^3$ (Ours) in Table~\ref{table:compare_to_baseline}.

As presented in Table \ref{table:compare_to_baseline}, our proposed method M$^3$ achieves the best overall performance across the three evaluated tasks. Importantly, M$^3$ attains these improvements without involving complex architectural designs or intensive computations; it merely adjusts the contribution ratios of constituent models to realize substantial gains, thereby outperforming the DELLA approach. DELLA advances model sparsification techniques by building upon TIES-Merging and DARE, introducing a novel pruning strategy termed MAGPRUNE, which selects delta parameters based on their magnitudes. This strategy has inspired us to investigate new directions for enhancing the performance of merged models.

\section{Performance Drop Rate (PDR) for Adversarial Robustness}
\label{sec:Performance Drop Rate (PDR) for Adversarial Robustness}
The adversarial robustness is evaluated using the Performance Drop Rate (PDR) \citep{zhu2023promptrobust}, which is defined as follows:
\begin{align}
\text{PDR} = \frac{\text{Metric}_{\text{no attack}} - \text{Metric}_{\text{attack}}}{\text{Metric}_{\text{no attack}}},
\end{align}
\noindent where $\text{Metric}_{\text{no attack}}$ denotes the performance metric without any prompt attack, and $\text{Metric}_{\text{attack}}$ represents the performance metric under the prompt attack. A smaller PDR indicates stronger adversarial defense against prompt attacks, implying better adversarial robustness.

\section{Additional Experimental Results on Adversarial Robustness}
\label{sec:Results on the Adversarial robustness}
\begin{table*}[t]
\centering
\small 
\begin{tabular}{>{\centering\arraybackslash}p{1.5cm} >{\centering\arraybackslash}p{1.5cm} >{\centering\arraybackslash}p{1.5cm} >{\centering\arraybackslash}p{1.5cm} >{\centering\arraybackslash}p{1.5cm} >{\centering\arraybackslash}p{2cm}} \toprule
Model & Dataset & Use Mixup & Use Attack & Metric (\%) & PDR (\%) \\ \hline

\multirow{8}{*}{\makecell{Math \\ \& Code}} & \multirow{4}{*}{SST2} & \multirow{2}{*}{No} & No  &  57.68 & \multirow{2}{*}{ 11.73} \\  
                                            &                       &                     & Yes &  50.92 &                         \\ \cline{3-6}
                                            &                       & \multirow{2}{*}{Yes}& No  &  78.21 & \multirow{2}{*}{ 37.10} \\  
                                            &                       &                     & Yes &  49.20 &                         \\ \cline{2-6}
                                            & \multirow{4}{*}{CoLA} & \multirow{2}{*}{No} & No  &  72.87 & \multirow{2}{*}{ 56.97} \\  
                                            &                       &                     & Yes &  31.35 &                         \\ \cline{3-6}
                                            &                       & \multirow{2}{*}{Yes}& No  &  74.02 & \multirow{2}{*}{ 58.94} \\  
                                            &                       &                     & Yes &  30.39 &                         \\ \cline{1-6}

\multirow{8}{*}{\makecell{LM \\ \& Math}} & \multirow{4}{*}{SST2} & \multirow{2}{*}{No}  & No  &  92.78 & \multirow{2}{*}{ 2.60} \\  
                                          &                       &                      & Yes &  90.37 &                         \\ \cline{3-6}
                                          &                       & \multirow{2}{*}{Yes} & No  &  91.28 & \multirow{2}{*}{ 3.77} \\  
                                          &                       &                      & Yes &  87.84 &                         \\ \cline{2-6}
                                          & \multirow{4}{*}{CoLA} & \multirow{2}{*}{No}  & No  &  79.19 & \multirow{2}{*}{ 4.96} \\  
                                          &                       &                      & Yes &  75.26 &                         \\ \cline{3-6}
                                          &                       & \multirow{2}{*}{Yes} & No  &  80.54 & \multirow{2}{*}{ 1.07} \\  
                                          &                       &                      & Yes &  79.67 &                         \\ \cline{1-6}

\multirow{8}{*}{\makecell{LM \\ \& Code}} & \multirow{4}{*}{SST2} & \multirow{2}{*}{No} & No  &  10.55 & \multirow{2}{*}{ 98.91} \\  
                                          &                       &                     & Yes &  0.11 &                         \\ \cline{3-6}
                                          &                       & \multirow{2}{*}{Yes}& No  &  73.17 & \multirow{2}{*}{ 97.65} \\  
                                          &                       &                     & Yes &  1.72 &                         \\ \cline{2-6}
                                          & \multirow{4}{*}{CoLA} & \multirow{2}{*}{No} & No  &  74.21 & \multirow{2}{*}{ 8.79} \\  
                                          &                       &                     & Yes &  67.69 &                         \\ \cline{3-6}
                                          &                       & \multirow{2}{*}{Yes}& No  &  73.922 & \multirow{2}{*}{ 11.15} \\  
                                          &                       &                     & Yes &  65.68 &                         \\ \bottomrule
\end{tabular}
\caption{\label{table:attack_results_deepwordbug} Adversarial robustness of merged models on the SST2 and CoLA datasets when executing the DeepWordBug prompt attack method.}
\end{table*}

\begin{table*}[t]
\centering
\small 
\begin{tabular}{>{\centering\arraybackslash}p{1.5cm} >{\centering\arraybackslash}p{1.5cm} >{\centering\arraybackslash}p{1.5cm} >{\centering\arraybackslash}p{1.5cm} >{\centering\arraybackslash}p{1.5cm} >{\centering\arraybackslash}p{2cm}} \toprule
Model & Dataset & Use Mixup & Use Attack & Metric (\%) & PDR (\%) \\ \hline

\multirow{8}{*}{\makecell{Math \\ \& Code}} & \multirow{4}{*}{SST2} & \multirow{2}{*}{No} & No  &  57.68 & \multirow{2}{*}{ 13.92} \\  
                                            &                       &                     & Yes &  49.66 &                         \\ \cline{3-6}
                                            &                       & \multirow{2}{*}{Yes}& No  &  78.21 & \multirow{2}{*}{ 4.11} \\  
                                            &                       &                     & Yes &  75.00 &                         \\ \cline{2-6}
                                            & \multirow{4}{*}{CoLA} & \multirow{2}{*}{No} & No  &  45.54 & \multirow{2}{*}{ 13.47} \\  
                                            &                       &                     & Yes &  39.41 &                         \\ \cline{3-6}
                                            &                       & \multirow{2}{*}{Yes}& No  &  71.72 & \multirow{2}{*}{ 17.25} \\  
                                            &                       &                     & Yes &  59.35 &                         \\ \cline{1-6}

\multirow{8}{*}{\makecell{LM \\ \& Math}} & \multirow{4}{*}{SST2} & \multirow{2}{*}{No}  & No  &  92.78 & \multirow{2}{*}{ 2.22} \\  
                                          &                       &                      & Yes &  90.71 &                         \\ \cline{3-6}
                                          &                       & \multirow{2}{*}{Yes} & No  &  91.28 & \multirow{2}{*}{ 0.0} \\  
                                          &                       &                      & Yes &  91.28 &                         \\ \cline{2-6}
                                          & \multirow{4}{*}{CoLA} & \multirow{2}{*}{No}  & No  &  79.19 & \multirow{2}{*}{ 12.00} \\  
                                          &                       &                      & Yes &  69.70 &                         \\ \cline{3-6}
                                          &                       & \multirow{2}{*}{Yes} & No  &  80.54 & \multirow{2}{*}{ 5.83} \\  
                                          &                       &                      & Yes &  75.84 &                         \\ \cline{1-6}

\multirow{8}{*}{\makecell{LM \\ \& Code}} & \multirow{4}{*}{SST2} & \multirow{2}{*}{No} & No  &  10.55 & \multirow{2}{*}{ 95.65} \\  
                                          &                       &                     & Yes &  0.46 &                         \\ \cline{3-6}
                                          &                       & \multirow{2}{*}{Yes}& No  &  73.17 & \multirow{2}{*}{ 55.02} \\  
                                          &                       &                     & Yes &  32.91 &                         \\ \cline{2-6}
                                          & \multirow{4}{*}{CoLA} & \multirow{2}{*}{No} & No  &  74.21 & \multirow{2}{*}{ 7.24} \\  
                                          &                       &                     & Yes &  68.84 &                         \\ \cline{3-6}
                                          &                       & \multirow{2}{*}{Yes}& No  &  73.92 & \multirow{2}{*}{ 7.52} \\  
                                          &                       &                     & Yes &  68.36 &                         \\ \bottomrule
\end{tabular}
\caption{\label{table:attack_results_bertattack} Adversarial robustness of merged models on the SST2 and CoLA datasets when executing the Bertattack prompt attack method.}
\end{table*}

All the merged models are obtained using the Task Arithmetic method.
Table~\ref{table:attack_results_deepwordbug} presents the detailed experimental results of the adversarial robustness of merged models on the SST2 and CoLA datasets applying the DeepWordBug prompt attack method.
Table~\ref{table:attack_results_bertattack} presents the detailed experimental results of the adversarial robustness of merged models on the SST2 and CoLA datasets applying the BERTAttack prompt attack method.

\section{Detailed Introduction to DARE}
\label{sec:Detailed Introduction to DARE}
DARE (Drop And REscale) \citep{yu2024language} is a model sparsification method designed to reduce the redundancy of delta parameters in fine-tuned models while preserving their task-specific capabilities. In SFT, model parameters are optimized to unlock abilities for specific tasks, with the difference between fine-tuned and pre-trained parameters referred to as delta parameters.

However, studies have shown that delta parameters are often highly redundant. DARE addresses this redundancy by randomly dropping a proportion \( p \) of delta parameters (referred to as the drop rate) and rescaling the remaining ones by a factor of \( 1/(1 - p) \). This simple yet effective approach enables DARE to eliminate up to 99\% of delta parameters with minimal impact on model performance, particularly in large-scale models, and it can be applied using only CPUs.

Beyond sparsification, DARE serves as a versatile plug-in for merging multiple homologous fine-tuned models (fine-tuned from the same base model) by reducing parameter interference. When combined with existing model merging techniques such as Average Merging, Task Arithmetic, and TIES-Merging, DARE facilitates the fusion of models while retaining or even enhancing task performance across multiple benchmarks. This effect is especially pronounced in decoder-based LMs, where DARE boosts task generalization. 

Experiments on AlpacaEval, GSM8K, and MBPP reveal that the merged LM has the potential to outperform any individual source LM, presenting a significant new discovery. Notably, the 7B model obtained through DARE merging, SuperMario v2, ranks first among models of the same scale on the Open LLM Leaderboard \citep{beeching2023open}. These improvements were achieved without the need for retraining, positioning DARE as an efficient and resource-friendly solution for model merging.


\section{Integrating M$^3$ into the TIES-Merging Model Merging Method}
\label{sec:Integrating M$^3$ into the TIES-Merging Model Merging Method}
Figure~\ref{fig:ties} shows the specific implementation approach to incorporating M$^3$ into TIES-Merging. After the steps of trimming parameters with lower magnitudes and resolving sign disagreements, the two models to be merged are denoted as \( M_1 \) and \( M_2 \). During the M$^3$ process, only the parameters that are preserved in both \( M_1 \) and \( M_2 \) are interpolated according to the model merging hyperparameter $\lambda_m$ to obtain the merged parameters. For parameters that are preserved in only one of the models, no interpolation is performed, and the original value from the preserved model is retained in the merged model.

\end{document}